\documentclass[10pt]{article}
\usepackage[top=0.85in,left=0.75in,footskip=0.75in,marginparwidth=2in]{geometry}

\usepackage[T2A,T1]{fontenc}
\usepackage[utf8]{inputenc}
\usepackage[russian,french,english]{babel}

\usepackage{amsfonts}
\usepackage{amsmath,amssymb,amsfonts}%
\usepackage{amsthm}%
\usepackage{mathrsfs}%

\usepackage{booktabs}
\usepackage{multirow}
\usepackage{enumerate}
\usepackage{url}
\usepackage[toc,page]{appendix}

\usepackage{nicefrac}
\usepackage{graphicx}
\usepackage{subfig}
\usepackage{hyperref}
\usepackage{url}            
\usepackage{booktabs}  
\usepackage{nicefrac}
\usepackage{microtype}

\usepackage{graphicx}
\usepackage{subfig}

\usepackage{nameref,hyperref}

\usepackage[right]{lineno}

\usepackage{microtype}
\DisableLigatures[f]{encoding = *, family = * }

\usepackage{setspace} 

\usepackage{changepage}

\usepackage[aboveskip=1pt,labelfont=bf,labelsep=period,singlelinecheck=off]{caption}

\usepackage{lastpage,fancyhdr,graphicx}
\usepackage{epstopdf}
\fancyheadoffset[L]{2.25in}
\fancyfootoffset[L]{2.25in}

\usepackage{color}

\definecolor{Gray}{gray}{.25}

\usepackage{graphicx}

\usepackage{wrapfig}
\usepackage[pscoord]{eso-pic}
\usepackage[fulladjust]{marginnote}
\reversemarginpar

\usepackage[
]{biblatex}
\begin{document}
\vspace*{0.35in}

\begin{flushleft}
{\Huge
\textbf\newline{Towards hypergraph cognitive networks as feature-rich models of knowledge}
}

\vspace*{0.15in}

{\Large
Salvatore Citraro\textsuperscript{1}\textsuperscript{*},
Simon De Deyne\textsuperscript{2},
Massimo Stella\textsuperscript{3}\textsuperscript{+},
Giulio Rossetti\textsuperscript{1}\textsuperscript{+},
}

\bigskip
\bf{1} KDD-Lab, ISTI (CNR), Pisa
\\
\bf{2} Computational Cognitive Science Lab, University of Melbourne
\\
\bf{3} Department of Psychology and Cognitive Science, University of Trento
\\ 
\bigskip
\bf{*} salvatore.citraro@phd.unipi.it
\\
\bf{+} Equal contribution as last authors

\end{flushleft}

\section*{Abstract}

Conceptual associations influence how human memory is structured: Cognitive research indicates that similar concepts tend to be recalled one after another. Semantic network accounts provide a useful tool to understand how related concepts are retrieved from memory. However, most current network approaches use pairwise links to represent memory recall patterns (e.g. reading "airplane" makes one think of "air" and "pollution", and this is represented by links "airplane"-"air" and "airplane"-"pollution"). Pairwise connections neglect higher-order associations, i.e. relationships between more than two concepts at a time. These higher-order interactions might covariate with (and thus contain information about) how similar concepts are along psycholinguistic dimensions like arousal, valence, familiarity, gender and others. We overcome these limits by introducing feature-rich cognitive hypergraphs as quantitative models of human memory where: (i) concepts recalled together can all engage in hyperlinks involving also more than two concepts at once (cognitive hypergraph aspect), and (ii) each concept is endowed with a vector of psycholinguistic features (feature-rich aspect). We build hypergraphs from word association data and use evaluation methods from machine learning features to predict concept concreteness. Since  concepts with similar concreteness tend to cluster together in human memory, we  expect to be able to leverage this structure. Using word association data  from the Small World of Words dataset, we compared a pairwise network and a hypergraph with N=3586 concepts/nodes. Interpretable artificial intelligence models trained on (1) psycholinguistic features only, (2) pairwise-based feature aggregations, and on (3) hypergraph-based aggregations show significant differences between pairwise and hypergraph links. Specifically, our results show that higher-order and feature-rich hypergraph models contain richer information than pairwise networks leading to improved prediction of word concreteness. The relation with previous studies about conceptual clustering and compartmentalisation in associative knowledge and human memory are discussed.

\doublespacing

\section{Introduction}

Words in language bear implicit, unexpressed features \cite{aitchison2012words}. When reading ``the pen is on the table'', we immediately consider ``pen'' as a concrete object, even though the sentence does not convey specific quantitative information about it \cite{montefinese2019semantic}. We think of ``building'' as something with a large size, of ``love'' as something abstract, of ``crime'' as something negative \cite{scott2019glasgow}. These features contribute to making human language complex and nuanced just as much as its cognitive reflection in the human mind \cite{aitchison2012words}. Theoretical models \cite{doczi2019overview,aitchison2012words,vitevitch2022can,castro2020contributions} informed by considerable experimental evidence \cite{vitevitch2021phonological,wulff2022using,valba2022k,zock2010deliberate} point out that linguistic knowledge is organised in an associative way, with ideas sharing many features being more tightly connected and easier to be acquired, processed and recalled one after another. The cognitive system apt at processing knowledge expressible with language is commonly called ``mental lexicon'' \cite{aitchison2012words,zock2010deliberate}. Differently from common dictionaries, the mental lexicon includes not only knowledge relative to meanings but also other phonological \cite{vitevitch2022can}, emotional \cite{de2021visual} and visual \cite{kennington2021enriching}  aspects of conceptual knowledge, among many other features \cite{scott2019glasgow,montefinese2019semantic}.

Quantitative investigations of the mental lexicon, its structure and functioning, have recently benefited from the advent of Big Data and network science \cite{siew2019cognitive,castro2020contributions}. Massive psycholinguistic experiments mapped thousands of concepts across multiple dimensions, providing quantitative estimates for word concreteness, imageability, valence/sentiment, arousal and many other features  (cf. \cite{scott2019glasgow}). Access to this big data fostered the creation of several large-scale network models, with thousands of nodes, representing knowledge in the mental lexicon as engaging in different types of conceptual associations \cite{siew2019cognitive,stella2017multiplex,de2013better}. Feature-sharing networks (which are different from feature-rich networks \cite{citraro2020identifying}) link concepts based on overlap in semantic features \cite{steyvers2005large}, or overlap in sounds, in the case of phonological networks \cite{vitevitch2022can}, or concept similarity in the case of synonymy networks \cite{steyvers2005large}, among many other possibilities. The proliferation of network representations, backed up by psychological theory, saw even more refined attempts at directly mapping memory recall patterns from the mental lexicon: Free associations map cue-target responses from memory, devoid of any specific semantic or phonological constraint affecting them \cite{de2019small,de2013better,kenett2014investigating}. Reading the prompt/cue ``book'' and immediately thinking of "chapter" creates a free association link ``book'' - ``chapter''. Continued free associations extend this task to consider up to three recalls \cite{de2013better}, e.g. reading ``math'' elicits ``bad'', ``hard'', ``wrong'' in an individual \cite{stella2019forma}. Modelling continued free associations as three cue-response links led to the creation of free association networks better suited to capture weak associations compared to single-response procedures as \cite{de2013better,de2019small}). 

From a knowledge modelling perspective, free association networks have been a valid approach to capture semantic cognition broadly, as previous work demonstrates they capture semantic relatedness between concepts \cite{kenett2017semantic}, differentiate individuals based on creativity  \cite{stella2019viability},  reflect the affective (positive/negative) connotations of concepts \cite{vankrunkelsven2018predicting,stella2019forma}. This has a range of applications as well. For example, word associations can be used to infer psychometric measures of mental distress in healthy populations \cite{fatima2021dasentimental}. The Small World of Words is a multilingual international research project on free associations, gathering millions of free associations across 17 languages \cite{de2019small}. Until now, these associations have been modelled as pairwise relationships between words. More in detail, by construction, the recall of free associates always takes place in relationship with the same underlying stimulus \cite{de2013better}. Considering only pairwise relationships between the cue and its responses led to networks explaining the most variance in several lexical tasks (for details see \cite{de2013better}). Adding also pairwise relationships between responses themselves was shown to deteriorate network performance in explaining variance within lexical tasks and also added noise in the form of weak memory recall patterns between related responses \cite{de2013better}. In order to overcome noise, other techniques of pairwise network filtering, like maximal planar graph embeddings or minimum spanning trees, have been successfully applied to free association networks (see \cite{siew2019cognitive,kenett2017semantic}). 
However, more work is needed to evaluate and understand the appropriateness of different network filtering technique \cite{zock2010deliberate}. 
Returning to the cognitive interpretation of one cue producing some activation signal stimulating recall of all responses at the same time \cite{de2013better,vankrunkelsven2018predicting}, thus giving rise to a higher-order interaction, we hereby propose a novel theoretical framework for modelling free association data: Cognitive hypergraphs.

Hypergraphs are complex networks where sets of nodes engage in the same (hyper)link simultaneously \cite{berge1984hypergraphs,battiston2021physics,rosas2022disentangling}. Whereas pairwise complex networks consider only links between two nodes, hypergraphs can consider connections among 3, 4 or more entities. In this way, hypergraphs can naturally encode for interactions between nodes of order higher than 2. This is strongly appealing for modelling free association data, as it enables for cue and responses to be combined together at the hyperlink level. The mathematics of hypergraphs originates from graph theory and combinatorics, with seminal work over graph isomorphism completed almost 40 years ago \cite{berge1984hypergraphs}. Only recently the formalism was extended by physicists and computer scientists to model a plethora of real-world complex systems \cite{battiston2022higher,battiston2020networks}. Marinazzo and colleagues used hypergraphs of information-theoretic associations between items in psychometric scales to reduce the impact of redundant information on identifying clusters of co-occurring symptoms compared to pairwise networks \cite{marinazzo2022information}. De Arruda and colleagues showed that analogous social contagion models on hypergraphs and pairwise networks would exhibit crucially different dynamics, with hypergraphs supporting critical phase transitions closer to empirical estimates and not reproduced by pairwise network structures \cite{de2020social}. Veldt and colleagues defined an affinity score for estimating homophily in groups, showing that in a scenario with 2 labels and equally sized hyperedges majority homophily can not be reached by both groups for a combinatorial impossibility of hypergraphs \cite{veldt2023combinatorial}. Sarker and colleagues extend the previous affinity score for groups with more than 2 labels and for simplicial complexes \cite{sarker2023generalizing}.
These examples are part of a quick multidisciplinary growth of data science models based on hypergraphs, which, however, contains a gap: Even comprehensive reviews of the field  \cite{battiston2022higher,battiston2021physics} currently lack cognitive case studies. 

To the best of our knowledge, our cognitive hypergraph framework represents a first-of-its-kind approach to modelling human memory and the mental lexicon through higher-order interactions \cite{siew2019cognitive} where concepts are represented as feature-rich nodes, i.e. nodes are endowed with vectors of psycholinguistic features \cite{citraro2022feature}. The framework introduced here thus contains two points of novelty: (i) it combines response-response and cue-response  beyond pairwise links through the mathematical formalism of hyperlinks; (ii) it enriches nodes with psycholinguistic features as to explore any interplay between higher-order interactions and conceptual features. 

Focusing on sets of freely associated targets and cues as hyperlinks and including feature-rich representations of concepts/nodes, we explore and quantify the predictive power of cognitive hypergraphs against pairwise networks and standard psycholinguistic norms (neglecting any network structure) in reproducing word-level features. To do so we first extracted  the +12,000 cue words from Small World of Words (SWOW) \cite{de2019small}. Next we determined the overlap with the words in the Glasgow lexico-semantic norms \cite{scott2019glasgow}. The resulting network consisted of cue-response pairs from SWOW  for 3586 nodes. Each node was characterized by 11 features (i.e. covariates in psycholinguistic terms) representing linguistic and psycholinguistic dimensions, namely valence, arousal, dominance, semantic size, concreteness, gender association, age of acquisition, familiarity, frequency, polysemy and length (see the Methods for descriptions of each). Within an interpretable machine learning framework, we aim to use either network or psycholinguistic features (or a combination of both), to predict a target covariate/feature of nodes. Emphasis is then given to comparing pairwise network features against hypergraph features or unstructured psycholinguistic norms. Interpretability \cite{kumar2020problems} stems from the development of trained artificial intelligence (AI) models where the influence of one feature on model performance can be quantified and interpreted directionally (e.g. a higher feature improves regression performance). In this work, we focused on word concreteness as the predicted variable \cite{montefinese2019semantic}, using all others as predictor variables. We put emphasis over concreteness since it represents a crucial latent feature of words (not measurable directly like frequency or length \cite{brysbaert2014concreteness}) that is vastly studied in cognitive neuroscience \cite{fliessbach2006effect} and has been shown to affect several aspects of semantic cognition from lexical processing to information retention and knowledge internalisation \cite{montefinese2019semantic}.  

We provide new quantitative evidence that cognitive hypergraphs outperform both psycholinguistic baseline models and pairwise networks in predicting word concreteness from free association data. Our results underline the potential of going beyond pairwise interactions for modelling associative knowledge in human memory. 

\section{Results}

We frame our analysis in the context of the studies about assortative mixing in the mental lexicon \cite{siew2013community, van2015examining, citraro2020identifying, citraro2022feature}. Assortative mixing is an emerging behavior observed in many systems, such that nodes with similar features tend to connect together and stay apart from nodes with dissimilar features: The most common example refers to social networks, where individuals are more likely to interact in social circles if they share common features such as age, political, leaning, etc \cite{mcpherson2001birds, newman2003mixing}.
Several studies propose a clustered mental lexicon such that groups of similarly concrete words would act as the building blocks of many cognitive processes, e.g., the formation of cue-response homogeneous patterns in memory recall \cite{van2015examining}. Therefore, it would be possible to use the aggregated information provided by such groups to reconstruct/predict words' own traits, i.e., the empirical ground truth values according to a psycholinguist norm.
For example, the concreteness of a word like ``caterpillar" (i.e., its empirical ground truth value) would be determined by words connected to it (``butterfly", ``cabbage", etc). In the following, we discuss the rationale behind the adoption of several graph- and hypergraph-based representations for word associations (\ref{subsec:rationale}), guided by psycholinguistic sources such as the Small World of Words (SWoW) project \cite{de2019small} and the Glasgow Norms \cite{scott2019glasgow} (\ref{subsec:setting}); finally, we discuss our main findings, namely that hypergraph-based modules of word associations overcome the other representations in the concreteness prediction task (\ref{subsec:prediction}).

 \begin{figure}[t!]
 \centering
  {\includegraphics[scale=0.35]{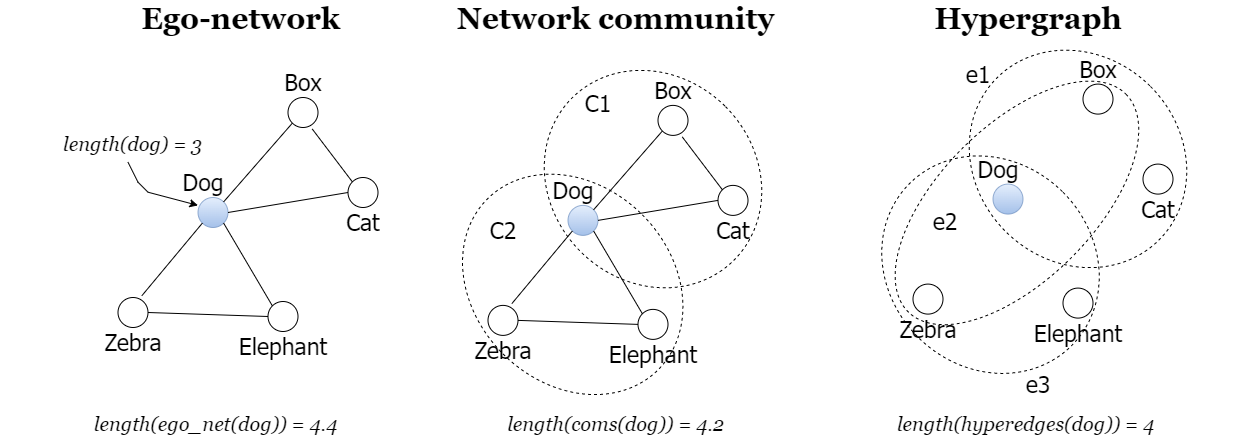}}
  \caption{A toy example showing different structural contexts surrounding the target word \textit{dog} in a network of free associations \cite{de2013better}.}
  \label{fig:models}
      \end{figure}

\subsection{Rationale of aggregation strategies}
\label{subsec:rationale}

\noindent \textbf{Ego-Networks}. Figure \ref{fig:models} describes several word labeling procedures, i.e., the expression of a module/context by means of a characteristic value.
We refer to a characteristic value of a context as the value associated to a target word as if that word was expressed by its direct (e.g. words directly linked) or indirect neighbors (e.g. words in the same community) rather than the word's own value.
The example in the figure is based on the aggregation of one single feature, \textit{length}, for one target word, \textit{dog}.
In Figure \ref{fig:models} (left), we leverage the ego-network of the free association network by just computing the average value of the feature, \textit{length}, in the neighborhood of the target word, \textit{dog}.
In this way, the length of the target word will be $4.4$ rather than $3$ (as if the word was expressed by the ego-network context), being the former one the average of the word set context \textit{box, cat, zebra, elephant} plus the target word itself, \textit{dog}, included. The reason why we include the target word as well in the context-set is because the target word is an essential constituent of the semantic/conceptual context. Removing the target word from its own context would create a gap/hole in the structure itself that could model/imply undesirable or partial knowledge (cf. Appendix A), e.g. without the star centre an ego network would just be a collection of disconnected components. Importantly, the addition of the target word contributes only to the creation of an aggregate measure, influenced by indirect/direct neighbors and their properties (as contrasting with the properties of the target itself).

\noindent \textbf{Contexts as local communities}. The aggregation based on the average value of nodes' ego-network is well-known and accepted in the literature of machine learning on graphs \cite{bhagat2011node}.
However, while reasoning about aggregation strategies in cognitive networks, one should consider that a word can be part of different contexts or neighborhoods \cite{firth1957synopsis, lenci2018distributional}. Hence, considering the whole ego-network could be an unsuitable proxy to estimate the value of a word \textit{by the company it keeps} \cite{firth1957synopsis}.
The free association network can still be used to identify more fine-grained contexts, e.g., the local communities surrounding a word \cite{fortunato2016community, li2015uncovering}.
Figure \ref{fig:models} (center) shows a toy partition centered around the word \textit{dog}.
The free association graph structure unveils that the target word can participate in two different contexts/communities, $C1 = \{dog, box, cat\}$, and $C2 = \{dog, zebra, elephant\}$.
This way the characteristic \textit{length} value in \textit{dog}'s context becomes the average of all the local communities/contexts where the word participates, $4.2$.

\noindent \textbf{Contexts as hyperedges}. However, contexts identified by ego-networks or network communities depend on an underlying network structure as a result of a heuristic process \cite{de2019small, zemla2020snafu}.
Hence, we leverage the expressive power of hypergraphs to induce a higher-order context from the participant responses. Rather than creating several pairwise links between a cue and its responses, the hyperedges of a hypergraph can connect multiple elements simultaneously \cite{battiston2021physics}. For each instance of the free association game, we model a hyperedge as the set that includes the cue word and all its responses. A response is thus modeled by means of a single connection rather than multiple pairwise links.

The characteristic value of a target word is calcuated as the average of the characteristic values of the hyperedges where the target word contributes in constituting an association pattern. 
In other words, while aggregating, we consider the so-called \textit{star ego-network} of a target word in a hypergraph: from \cite{comrie2021hypergraph}, the star ego-network of a node $u$ in a hypergraph is defined as the set of all the hyperedges that include $u$. For the sake of simplicity, we do not consider here other connections among the connected hyperedges, as in other fine-grained definitions of higher-order ego-networks \cite{comrie2021hypergraph}. Let us discuss a brief example of the star ego-network.

Figure \ref{fig:models} (right) shows a set of responses involving the word \textit{dog}.
Three possible outcomes, i.e., hyperedges, indeed are $e1=\{dog, box, cat\}$, $e2=\{zebra, dog, box\}$, and $e3=\{dog, zebra, elephant\}$.
Word associations here are not constrained to pairwise relations only.
For instance, in the toy association network there is no any direct link between \textit{zebra} and \textit{box}.
This could happen for several reasons depending on the strategy used for reconstructing the graph.
A possible explanation could be the following one. In the response \textit{zebra, dog, box}, \textit{zebra} is the cue word, \textit{dog} is the first and \textit{box} is the second response came to mind to the participant. Using a graph construction strategy where only consecutive words are connected, like a chain \cite{de2019small}, \textit{zebra} is not directly connected to \textit{box}, but only indirectly connected through \textit{dog}.
Conversely, the hypergraph model merges all the three words by means of a single hyperedge.
Doing so, the characteristic \textit{length} value in \textit{dog}'s context is not an average of all the graph-based contexts where the word participate ($4.4$, or $4.2$) but an average of all its higher-order contexts, $4$.

\subsection{Setting the stage}
\label{subsec:setting}

\begin{figure}[t!]
\centering
{\includegraphics[scale=0.30]{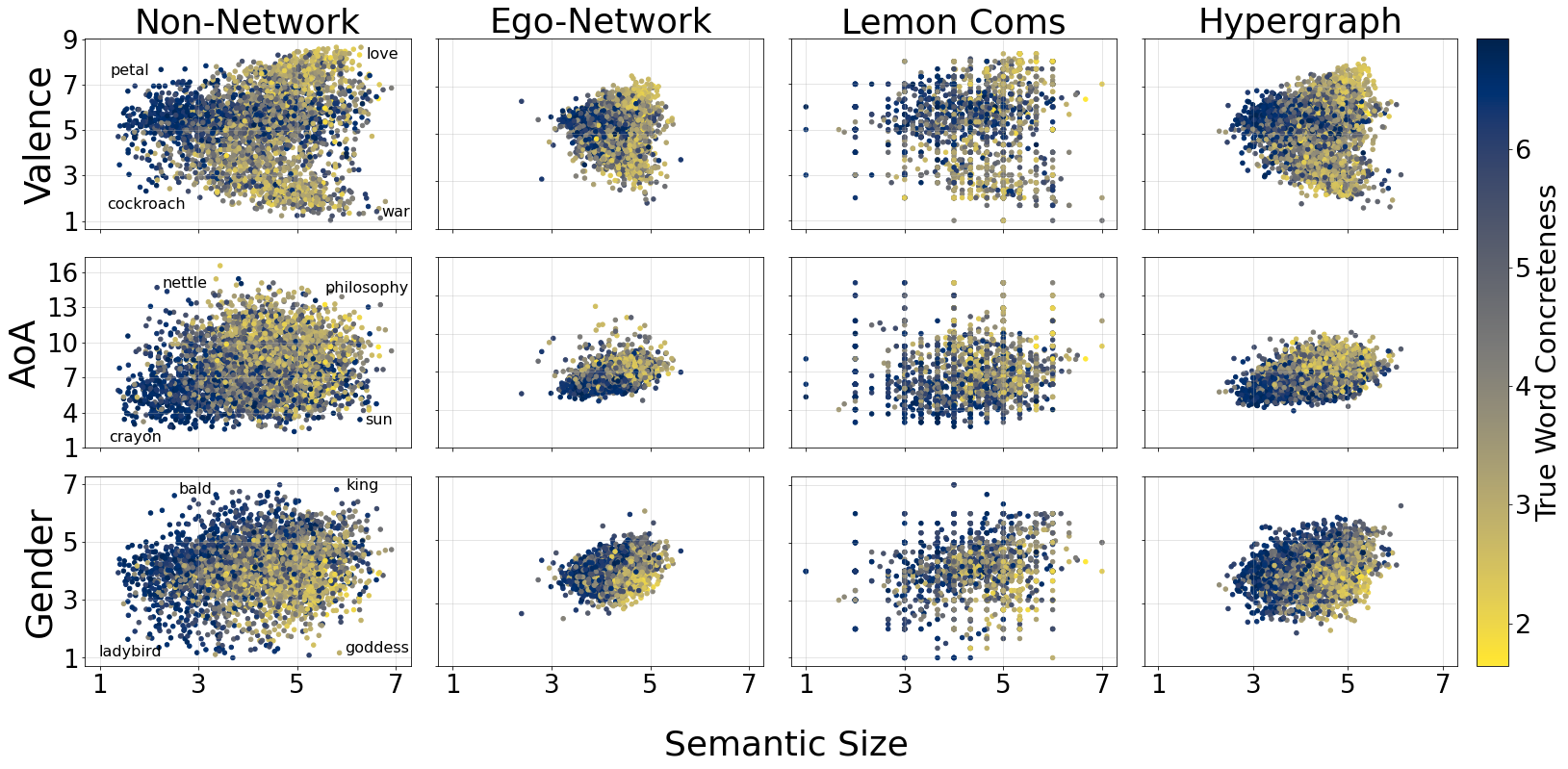}}
\caption{Scatter plots between the most important features according to the SHAP-values explanation (cf. Figure \ref{fig:shap_rf}). Each column represents an aggregation strategy, except for the first one. Points are always colored according to the original Glasgow Norms' concreteness.}
   \label{fig:scatter_concr_a}
      \end{figure}

\noindent \textbf{Data overview}. We gain patterns for 3586 English words present both in the SWoW \cite{de2019small} and in the Glasgow norms \cite{scott2019glasgow} projects.
From SWoW, we build the underlying graph/hypergraph structure; from the Glasgow Norms and other linguistic information easily available from words we form the vector of features to aggregate (cf. Section \ref{sec:mat_meth}). Figure \ref{fig:scatter_concr_a} provides a coarse-grained picture of the patterns emerging from different strategies. 
Each column provides an aggregation strategy. Each plot provides the characteristic values, except for the first one, where each point describes the empirical ground truth value in the Glasgow Norms, e.g., \textit{love} is an abstract (low concreteness) and salient (high semantic size) word associated to very positive emotions (high valence).
In the second column, based on the ego-network strategy, the characteristic values result in a more flattened, overall compact cloud of points.
Conversely, the hypergraph-based strategy comes as a hybrid between the non-network and the ego-network characteristics values, while the network community average values provide more coarse-grained value distributions (cf. later, Lemon communities).

\noindent \textbf{Outline of aggregation algorithms}. Here is our methodology to extract/aggregate word features:

\begin{itemize}
    \item \textit{Non-Network}: No aggregation strategies are defined, i.e., we do not use any underlying structure from SWoW to extract a characteristic value;
    \item \textit{Ego-Network}: Each word is described by a set of features whose characteristic value is the average of the word's ego-network (cf. \ref{subsec:rationale});
    \item \textit{Network communities}: We use different community-based strategies for feature aggregation; communities are found by using (i) a non-overlapping connectivity-based \cite{blondel2008fast} community detection algorithm; (ii) a non-overlapping both connectivity- and feature homogeneity-based \cite{citraro2020identifying} algorithm; (iii) an overlapping local expansion method \cite{li2015uncovering}; in detail:
    \begin{enumerate}[i]
        \item \textit{Louvain} \cite{blondel2008fast}: Same strategy as word's ego-network for aggregation. However, crisp communities provide larger contexts than ego-networks, since communities can group also nodes that are not directly neighbours \cite{fortunato2016community}. The Louvain method is based on the family of algorithms that optimize the modularity function;
        
        \item \textit{Louvain ``E''xtended to ``V''ertex ``A''ttributes (EVA)} \cite{citraro2020identifying}: Same strategy as word's ego-network for aggregation. EVA is an extension of Louvain that optimizes a linear combination of modularity and purity, a homogeneity-aware fitness function. Feature homogeneity-aware algorithms such as EVA force aggregations between words sharing similar feature values, in accordance with the word feature-homogeneity hypothesis \cite{van2015examining,citraro2022feature};
        
        \item \textit{Lemon} \cite{li2015uncovering}: This strategy labels each target node with the average value of the local average context of the target word (cf. \ref{subsec:rationale}). The algorithm can capture small sets of overlapping communities. Rather than identifying a crisp/global structure, \textit{Lemon} detects local modules given a representative set of seed nodes (cf. Materials and Methods). We run the algorithm $N$ times, where in each run the seed node is a different word; this way, we can detect the local communities centered around all target words. 
    \end{enumerate}
    
    \item \textit{Hypergraph}: This strategy labels each target node with the average value of the hypergraph-based characteristic value contexts of the target word (cf. \ref{subsec:rationale}).
\end{itemize}

\smallskip

\noindent \textbf{Details on prediction}. We test different algorithms from different families of methods to predict the concreteness value of a node.

\begin{itemize}
    \item \textit{Multiple Linear regression} \cite{fisher1922goodness}: Concreteness is expected to be a linear combination of the set of independent variables. The objective is to minimize the residual sum of squares between the observed targets (i.e., the original concreteness values) and the target predicted by the linear approximation;
    \item \textit{Random Forest} \cite{breiman2001random}: Several decision trees are built and the final output is based on the average of their predictions;
    \item \textit{AdaBoost} \cite{freund1997decision, schapire2013explaining}: An ensemble method where a combination of weak estimators, e.g., decision stumps, are built sequentially to produce a stronger output; 
    \item \textit{Support Vector Machine} \cite{platt1999probabilistic}: SVM's are used to find an appropriate hyperplane to fit the data while trying to define how much error is acceptable in the model.
\end{itemize}

The algorithms provided similar results both in terms of evaluation performances and model explanation. We show in the main article the one that outperformed the others, the Random Forest (cf. Appendix B).
Note that for each algorithm we provide hyperparameter tuning to maximize performances, and all the performance evaluations are cross-validated (cf. Section \ref{sec:mat_meth}).

\subsection{Predicting concreteness}
\label{subsec:prediction}

\begin{figure}[t!]
\centering
  \subfloat[RMSE]{\includegraphics[scale=0.35]{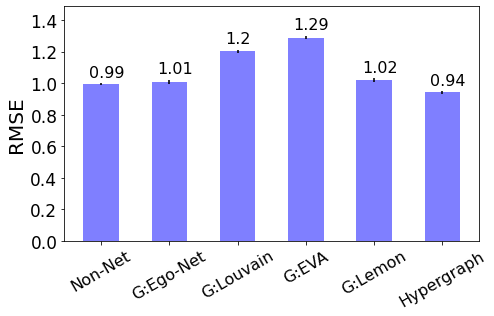}}
  \subfloat[$R^2$]{\includegraphics[scale=0.35]{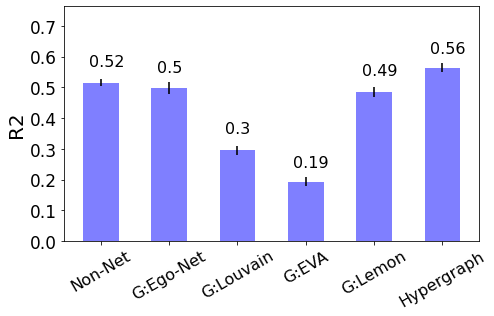}}
  \caption{Random Forest evaluation of concreteness prediction based on the different aggregation strategies.}
  \label{fig:eval_rf}
      \end{figure}

We present here the Random Forest (henceforth, RF) performances on each dataset (cf. Section \ref{sec:mat_meth} for the RF hyperparameter tuning and Appendix B for other methods).
The evaluation metrics in Figure \ref{fig:eval_rf} highlight theperformances in terms of the average distance between predicted and original values, i.e., using the Root-Mean Squared Error (RMSE), and the variation in the variable in percentage terms ($R^2$).
See \textit{Materials and Methods, Evaluation details} for a precise description of the formulas.
As can be seen from Figure \ref{fig:eval_rf}, the RF regressor provides better predictions on the set of features based on the hypergraph aggregation, while all the community-based strategies make the RF perform worse; performances on the ego-network aggregation and on the non-network strategy are similar.

\begin{figure}[t!]
  \centering
  \subfloat[Non-Network]{\includegraphics[scale=0.33]{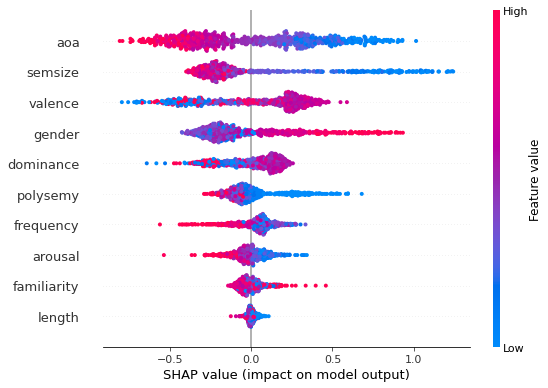}}
  \subfloat[Ego-Network]{\includegraphics[scale=0.33]{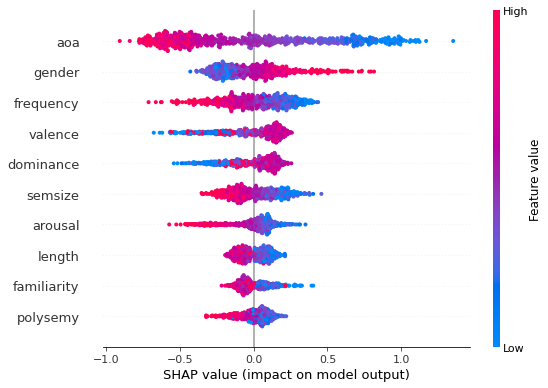}}
  \qquad
  \subfloat[Lemon coms.]{\includegraphics[scale=0.33]{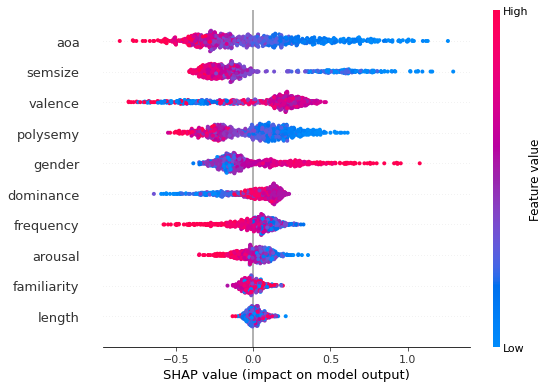}}
  \subfloat[Hypergraph]{\includegraphics[scale=0.33]{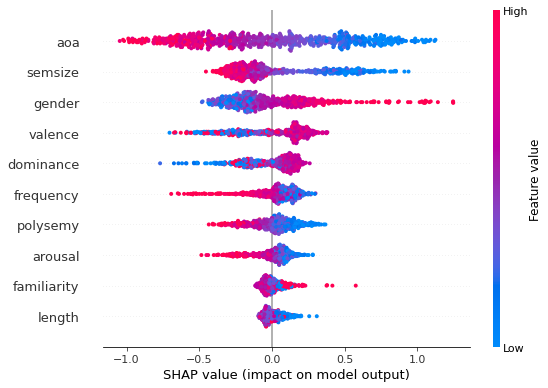}}
  \caption{Random Forest feature importance based on SHAP-values. Features ordered according to their importance.}
  \label{fig:shap_rf}
      \end{figure}

Figure \ref{fig:shap_rf} presents a more fine-grained evaluation based on feature importance with SHAP values \cite{NIPS2017_7062, lundberg2020local2global}.
This two-fold evaluation highlights (i) which set of features provides better information in the estimation of word concreteness (RMSE, $R^2$); (ii) which features and feature values are useful to the model in the prediction. Different regression techniques are evaluated in Appendix B.
Figure \ref{fig:shap_rf} combined with Figure \ref{fig:eval_rf} show that the RF achieves better predictions (on almost all the sets of features, net of different performances) when values of age of acquisition and semantic size are low, and when the values of valence are high, as well as when words are associated to a masculine aspect of salience (high values of the gender variable).

To understand what these word profiles mean and how they can provide useful aggregated patterns, let us focus again on Figure \ref{fig:scatter_concr_a}.
The scatter plots tell us there is correlation between concreteness and some other variables like valence, age of acquisition, gender and semantic size.
For instance, there is a consistent group of early acquired, masculine-associated, concrete words with low values of semantic size and high values of valence.
Also, there are some abstract words, i.e., words with low concreteness values, which are associated with medium-high values of semantic size.
In fact, semantic size can be thought of as a proxy for conceptual salience across both abstract and concrete words, thus correlation with both concrete and abstract words is expected \cite{yao2013semantic}.
See \textit{love} and \textit{war}, for instance, which are two extremely high semantic salient words with opposite valence , where \textit{love} is highly abstract, and \textit{war} is highly concrete; cf. also \textit{philosophy/sun} and \textit{king/goddess} (cf. Figure \ref{fig:scatter_concr_a}).

\begin{figure}[t!]
  \centering
  \subfloat[]{\includegraphics[scale=0.29]{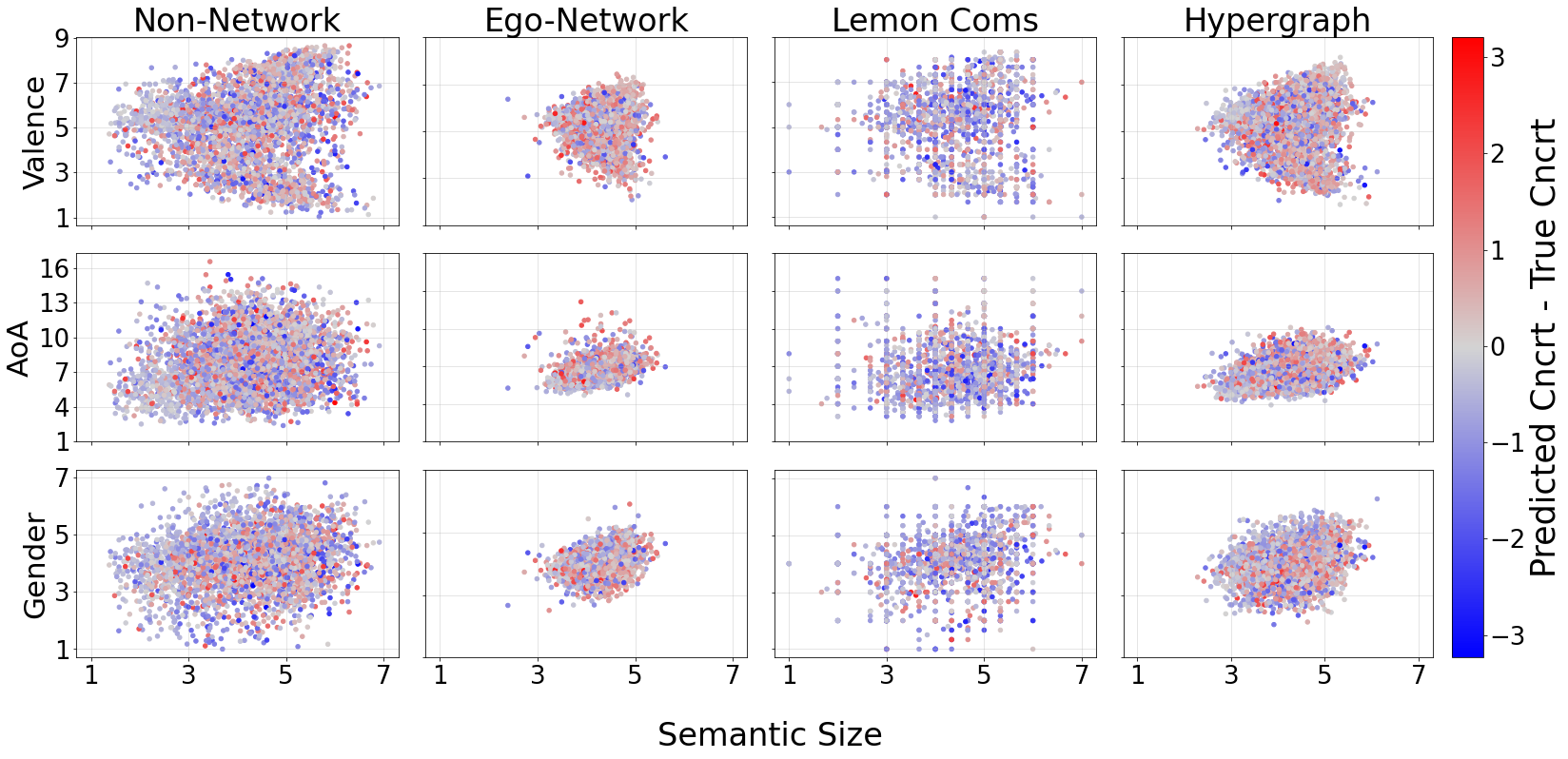}}
  \caption{Scatter plots between the most important features according to the SHAP-values explanation (cf. Figure \ref{fig:shap_rf}). Points are colored according to the difference between the value predicted by the RF model and the empirical ground truth value.}
  \label{fig:scatter_concr_b}
      \end{figure}

According to Figure \ref{fig:scatter_concr_a}, the correlation remains unchanged in all the aggregation strategies.
The combined results from Figure \ref{fig:shap_rf} and Figure \ref{fig:scatter_concr_b} highlight that the RF can well predict a set of high concrete words associated with some characteristics such as early word acquisition or positive emotion.
Figure \ref{fig:scatter_concr_b} complements feature importance (cf. Figure \ref{fig:shap_rf}) and scatter plots (cf. Figure \ref{fig:scatter_concr_a}) by coloring each word with respect to the residuals, i.e., the differences in the predicted and original concreteness.
Note the "grey" zones, that indicate the words for which such differences in the predicted and empirical ground truth values are small: in this way, we can verify that the RF predicts the values of concrete words with the previous mentioned characteristics, validating the impact given by the SHAP values to profiles as positive valence and early word acquisition (cf. Figure \ref{fig:shap_rf}).
From Figure \ref{fig:scatter_concr_b}, we can see that also abstract words can be well predicted by the RF; however, no clear patterns as the ones highlighted by SHAP summary plots emerge for the prediction of abstract words.
Finally, Figure \ref{fig:scatter_concr_b} shows no noticeable variations in residuals across the different strategies.
This indicates that the enhancement achieved through the utilization of hypergraph-based aggregation is attributable to improved regression (cf. Figure \ref{fig:eval_rf}) rather than the ability to predict specific profiles that cannot be captured by alternative aggregation methods or empirical ground truth values.

\section{Discussion}

Our work moves a step forward towards using hypergraphs \cite{battiston2022higher} in cognitive modelling: Using hypergraphs provides richer cognitive measures compared to techniques that rely on communities or local neighborhoods. In other words, we show that the hypergraph formalism is better than pairwise networks or unstructured sets of features at predicting concreteness norms for individual words. Regression models on unstructured features try to predict a psycholinguistic norm of a target word/concept based on the word's own values, neglecting any conceptual association the target might have with other concepts. Why would connectivity matter? Recent work in cognitive network science has highlighted how memory recall patterns like the ones captured here can be highly insightful about semantic relatedness \cite{kenett2017semantic,kumar2020distant}, indicating that words separated by fewer memory recalls (i.e. shortest path length in terms of free associations) tend also to be rated as more semantically related. Shorter distance on free association networks thus corresponds to higher semantic relatedness.

Our working hypothesis is that the proximity between nodes in a semantic network translates into analogous values for mostly semantic psycholinguistic features, like concreteness \cite{fliessbach2006effect}. Under this hypothesis, words closer to a target share similar concreteness norms and could thus enable quantitative predictions for the concreteness of the target itself. Consequently, our working hypothesis corresponds to the presence of a compartmentalisation of semantic features and network structure in the mental lexicon, where clusters of closer words can tend to share similar concreteness norms. Importantly, our work cannot identify a causal relationship, e.g. are the words connected because they are equally concrete, or are they concrete, because they have a certain number of connections? Despite this limit, our assumption identifies an insightful correlation. Network structure might thus be valuable for predicting the concreteness of one word by considering its close words/neighbours on a network topology of memory recall patterns. This hypothesis is supported by preliminary evidence in a previous work with pairwise network \cite{vankrunkelsven2018predicting}. We test  three ways for selecting neighbours to a given target word: (i) words linked to the target (i.e. network neighbourhood) based pairwise edges between cues and responses, (ii) words in the same community of the target in based on pairwise cue-response edges, and (iii) words linked to the target by sharing a hyperlink in a hypergraph representation of cue-response pairs. Notice that community analysis within the hypergraph representation of free associations \cite{battiston2022higher} found trivial communities, which were discarded from the comparison. 

We test our hypothesis through a machine learning framework. Model performance reports quantitative evidence that hyperlinks constitute the best proxy for predicting words' concreteness, outmatching both unstructured and structured models based on pairwise network neighbourhoods and communities. 

These results confirm our working hypothesis and quantitatively indicate the presence of compartmentalisation in the layout of word associations that emerges more prominently when hypergraphs, rather than pairwise links in association graphs are considered. This clustering might emerge more in hypergraphs because they do not impose any specific distinction between the cue (e.g. ``letter'') and the responses (e.g. ``mail'', ``sign'', ``dear''), which get represented within the same mathematical element (e.g. the hyperlink {``letter'',``mail'',``sign'',``dear''}). In pairwise networks, instead, the cue is automatically a more relevant node than its responses \cite{de2013better}, since the associations are encoded as links where the cue appears 3 times more frequently than the responses themselves, e.g. {(``letter'',``mail''),(``letter'',``sign''),(``letter'',``dear'')}. Not all words in free association networks are used as cues with the same frequency \cite{de2019small}, this dichotomy leads to structurally different networks, whose predictive power of concreteness norms is different. 

Cognitive hypergraphs represent a relatively novel tool for cognitive modelling because they are able to highlight a compartmentalisation phenomena that would be otherwise invisible with mainstream pairwise networks modelling free association data. Notice that we use the term ``compartmentalisation'' in a different way compared to previous approaches. In psychology, compartmentalisation is a strategy for separating conflicting and non-conflicting ideas \cite{ditzfeld2014self}. We rather use this notion to identify a tendency for associative knowledge in the mental lexicon to form networked clusters/compartments of words sharing similar concreteness rates and appearing as being hyperlinked together. Unlike taxonomic categories, which are made of words sharing a common theme (e.g. all words being "animals" \cite{de2015using}), compartments identify coherence in terms of a semantic psycholinguistic feature (e.g. all words being highly concrete). 

    \begin{figure}[t!]
    \centering
  \subfloat[]{\includegraphics[scale=0.18]{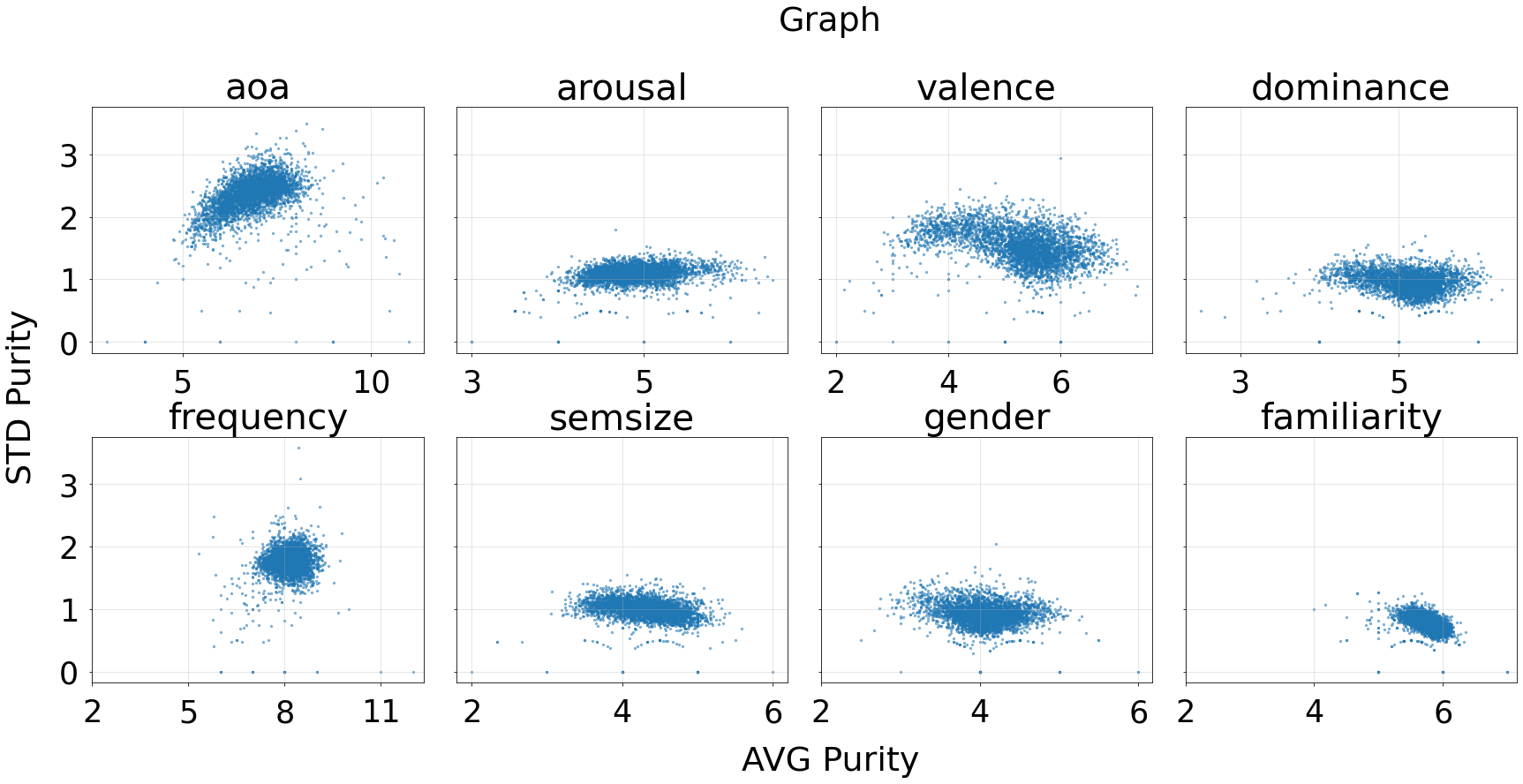}}
  \qquad
  \subfloat[]{\includegraphics[scale=0.18]{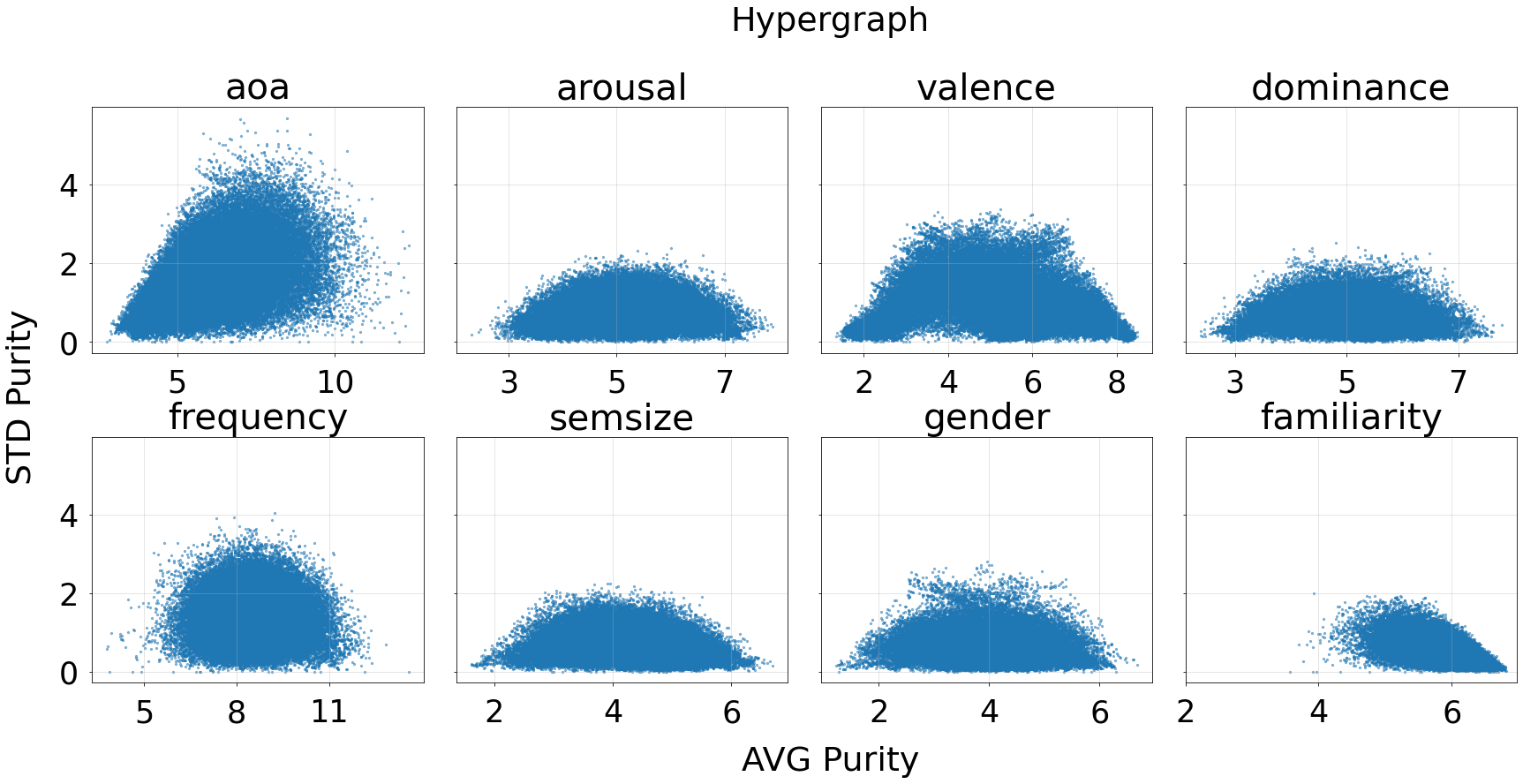}}
  \qquad
  \subfloat[]{\includegraphics[scale=0.18]{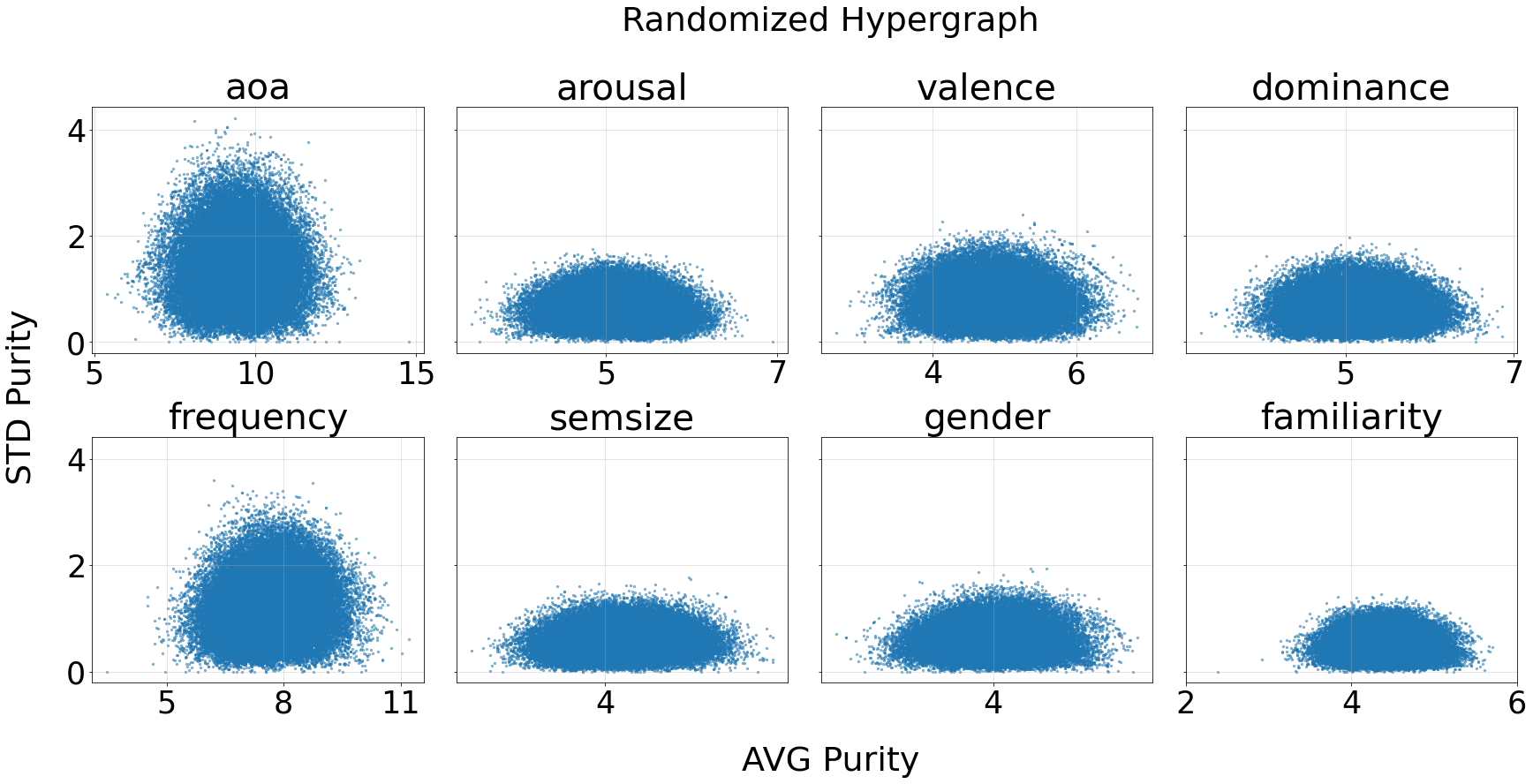}}
  \caption{Mean-standard deviation scatter plots of graph ego-network (a), hypergraph star ego-network purities (b) and its randomized representation (c) in all the dependent variables (polysemy not showed for better readability). }
  \label{fig:meanstd}
      \end{figure}

Our finding of feature-, hypergraph-based compartments in the mental lexicon agrees with previous works indicating a cognitive advantage in processing together more similar concepts \cite{kumar2020distant,hills2015foraging,todd2020foraging}. Compartments might reflect a tendency for associative knowledge to be sorted in ``patches'' of concepts being thematically non-coherent but still similar in terms of some psycholinguistic norms. In other words, compartments might reflect patterns of semantic foraging in the organisation and search of mental knowledge. Future research might investigate pre-existing frameworks for semantic foraging \cite{hills2015foraging,todd2020foraging} with novel contributions from hypergraphs. A challenge for this kind of research would be the assertion of which psycholinguistic features are mere consequences of more basic elements (e.g. frequency, length) and which are, instead, encoded properties of concepts, like concreteness, that cannot be fully explained by such basic elements only \cite{charbonnier2019predicting}.

Notice that non-semantic psycholinguistic features might not give rise to compartmentalisation. In our tests, predicting a not purely semantic norm like the age of acquisition (AoA) of words (which does not depend only on semantics but also on phonological and orthographic features of words \cite{brysbaert2000age,brysbaert2017test}) resulted in regression models of unstructured norms behaving way better ($R^2= 0.6 \pm 0.02$) than network-based pairwise ($R^2$ 0.25 ± 0.02) and hypergraph ($R^2 = 0.45 \pm 0.03$) models (cf. Appendix C). Furthermore, hypergraph models behaved worse than unstructured norms even when predicting arousal, dominance, familiarity and length. Nonetheless, hypergraph models behaved significantly better (at least 5 times better in terms $R^2$) than pairwise network model in predicting these other 5 psycholinguistic dimensions. These differences are expected, since our working hypothesis relies on the finding that network distance reflects mostly semantic similarity. Non-semantic aspects of words might be affected in other ways by network structure, thus decreasing the performance of network-based models in predicting non-purely semantic norms (like AoA). When considering pairwise network, we can offer an intuitive argument about this lack of predictive power rising from network patterns. Previous works have shown that in pairwise networks non-semantic features follow disassortative rather than assortative patterns. Affective patterns like valence were shown to make pairwise free association networks become disassortative \cite{van2015examining,stella2019forma}, i.e. pairwise links connected words with opposite sentiment/valence polarities which often occur as antonym pairs (\textit{pretty} - \textit{ugly}, \textit{young} - \textit{old}) in free association pairwise networks. Disassortativity made pairwise network models powerful predictors of words' sentiment/valence \cite{vankrunkelsven2018predicting}, a pattern that we here explored under the framework of cognitive hypergraphs as introduced here. Cognitive hypergraphs surpassed both unstructured norms and pairwise networks in predicting valence (cf. Appendix C). This finding indicates that although parwise disassortative patterns exist in the network encoding of memory recalls, there is a stronger tendency for valence coherence to persist in subsequent recalls. Similarly to the mechanism of compartmentalisation we outlined above, this valence coherence creates clusters of words with similar valence and it cannot be captured unless one considers higher-order interactions, going from pairwise to hypergraph formalisms. Our findings thus indicate that non-semantic compartmentalisations can be noticeable in psycholinguistic data and push for more data-informed explorations of the organisation of psycholinguistic features within networks of memory recall patterns.

Compartmentalisation is present not only across the hyperlinks in a given neighbourhood but also among words within a single hyperlink. This tendency is even more evident for extreme values of norms. For instance, in Figure \ref{fig:meanstd} (b), many hyperlinks tend to have words with similarly low age of acquisition norms. The extremes in Figure \ref{fig:meanstd} (b) are not a statistical artefact when they cannot be reproduced by randomly sorting words in hyperlinks, which is the case for Figure \ref{fig:meanstd} (c). This difference indicates a tendency for words in hyperlinks to be more similar in terms of age of acquisition, arousal, valence, dominance, semantic size, gender and familiarity when their average value for that norm is extreme, i.e. extremely low or high. This pattern further indicates a tendency for words to get compartmentalised even within hyperlinks and this might be due to an advantage in recalling concepts with similarly extreme psycholinguistic norms \cite{hills2015foraging}. 

It has to be noted that compartmentalisation between concepts was quantitatively captured also by parallel distributed processing (PDP) models \cite{farah1991computational,rogers2004semantic,}. PDP models quantify connections among individual features of each concept and then related knowledge retrieval to the strengths of the connections (e.g. the overlap in features) between elements \cite{shabahang2022generalization}. Despite this analogy, PDP models and cognitive hypergraphs adopt distinct representations of semantic memory. PDP models encode similarities in computational ways, so that concepts are related by means of a dynamical process or signal spreading across them \cite{farah1991computational}. Cognitive hypergraphs encode local relationships directly from empirical data, without needing additional computations. In this way, cognitive hypergraphs are more transparent than PDP models and can shed more light on the interplay between representational aspects of conceptual similarities and memory recalls, nonetheless PDP models can provide more insights about the dynamics of memory recall patterns and its failures \cite{schapiro2017complementary,farah1991computational}. Future research could potentially merge representational and dynamical aspects of both modelling approaches to investigate memory recalls more closely.

In terms of limitations, one of the most important ones is relative to filtering free associations in hypergraphs. Firstly, Glasgow norms represent one among many repositories for psycholinguistic norms, see \cite{brysbaert2014concreteness,brysbaert2017test,montefinese2019semantic}. Based on the positive pioneering findings gained from this study, future research could test larger repositories of psycholinguistic variables that cannot be directly encoded in terms of network structure. The South Carolina Psycholinguistic Metabase (SCOPE) \cite{gao2022scope}, which features 245 different lexical norms for 105,992 English words, represents a powerful candidate for future investigations with feature-rich hypergraphs, like the ones outlined here, and pairwise networks, like the ones investigated in \cite{citraro2022feature}.
Several prior works on free associations in pairwise networks have used some sort of filtering of infrequent or redundant word associations \cite{stella2019forma,kenett2017semantic}. Cognitive hypergraphs might not account for a statistical filtering of hyperlinks in some instances. In this dataset, applying the same statistical filtering introduced in \cite{musciotto2021detecting}, dismantled the whole set of hyperlinks. With link filtering being relevant for identifying meaningful network relationships and noisy links \cite{kenett2014investigating,siew2019cognitive}, more techniques should be tested and designed in cognitive modelling settings. Another limitation of our approach revolves around a black-box nature of machine learning models \cite{rudin2019stop}, which are not yet commonly used in psychology. Black-box models make it difficult for the experimenter to identify how data is internally represented within the model, e.g. feature X being higher promotes the prediction of outcome Y. We try to address this issue by using Shapley values \cite{ghorbani2019data}, a game-theoretic set of estimators for feature importance and contribution to model predictions. Although providing additional model interpretability, of relevance for cognitive modelling, Shapley values cannot provide causal evidence (feature X causes a better prediction of outcome Y) but only weaker correlation patterns \cite{kumar2020problems}. Despite this, Shapley values were crucial to identify compartmentalisation in our data and should thus be more commonly used in future investigations merging artificial intelligence and cognitive modelling. Last but not least, this first-of-its-own investigation of cognitive hypergraphs as psychological models is indeed limited by the modest amounts of behavioural effects being considered here, i.e. the modelling presented here explored only free association data whereas modelling the mental lexicon might encompass multiple layers of behavioural data \cite{citraro2022feature,stella2017multiplex}. This limitation is mainly due to the fact we focused our working hypothesis in terms of compartmentalisation within memory recalls only, without considering other psychological effects (e.g. reaction times in lexical decision-making tasks). Future works might explore whether the compartmentalisation found here could explain some variance in reaction times due to the dimensions that we found being well-captured by cognitive hypergraphs, i.e. concreteness and valence.

\section{Materials and Methods}
\label{sec:mat_meth}

\noindent \textbf{Free associations}. The Small World of Words (SWoW) project\footnote{\url{https://smallworldofwords.org/en/project/home}} \cite{de2019small} is a large-scale database that aims to build mental dictionaries/lexicons in different languages from a word association test  where each participant is asked to respond with at most 3 words coming to mind given a cue word.
In this study we use the English lexicon (SWOW-EN), although other datasets in Dutch and Spanish are also available and new languages will be added in the future \footnote{\url{https://smallworldofwords.org/en/project/stats}}.

\noindent \textbf{Features}. The Glasgow norms \cite{scott2019glasgow} provide a multidimensional set of psycholinguistic variables describing a word in terms of emotion conveyed (valence, dominance), salience (semantic size, arousal, gender association), exposure (age of acquisition, familiarity), and visualization (concreteness).
We use all the features available from this dataset except for age of acquisition, replaced with the data from \cite{kuperman2012age}, which provide more fine-grained information than the two-years binning from the Glasgow norm variable.
Moreover, to increase the number of word dimensions, we also add information about word length, frequency and polysemy degree.
Frequency is obtained from the OpenSubtitle dataset \cite{barbaresi2014language}, and polysemy values are proxied by the size of the WordNet synsets \cite{miller1995wordnet}.
A pre-processing step is needed before using the frequency variable, namely a logarithmic transformation, due to the well-known heavy-tailed distribution of this variable in human language \cite{zipf2016human}. Notice that when used for predictions, different variables are scaled to reduce normalisation issues.
\\
\noindent \textbf{Aggregation details}. For the creation of the free association network we strictly follow the \textit{R123} procedure described in \cite{de2019small}, namely that a link is formed between all the three responses and the cue word.
Note that the responses are not connected in their turn to each other.
The resulting graph $G=(V_G,E_G)$, with the filtering due to the matching between the SWoW and the Glasgow Norms words, has $V_G=3586$ and $E_G=165,690$.
See also Appendix D for other pairwise-based aggregation strategies and the resulting graphs.
The algorithms used for identifying communities depend on some parameters.
A standard and accepted value of the resolution limit parameter $\gamma$ is used for the Louvain algorithm, $\gamma=1$.
Moreover, the EVA algorithm, an attribute-aware extension of Louvain, also depends on a parameter $\alpha$, that tunes the importance of forcing homogeneity within communities (the higher, the more homogeneous communities are identified). We set $\alpha=0.8$ to obtain a partition significantly different from the Louvain one.
Lemon is an algorithm from the family of seed set expansion methods, that neglect the global structure for identifying local modules expanding from a set of seed nodes. Usually, the seeding strategies involve random walks aiming to optimize some fitness score for communities \cite{whang2013overlapping, kloumann2014community}. In detail, Lemon constructs the local spectra based on the singular vector approximations drawn from short random walks \cite{li2015uncovering}.
We use the original parameter values used in the Lemon algorithm paper \cite{li2015uncovering}, except for a preference on the maximum community size, set to $4$ to explicitly simulate the set size of the SWoW responses.

Finally, the hypergraph $H=(V_H,E_H)$ resulting from the intersection between the SWoW and the Glasgow norms vocabularies has $V_H=3586$ and $E_H=67,600$.
\\
\noindent \textbf{Prediction details}. In the RF model, we have chosen the best set of parameter values for the number of estimators (number of trees in the forest), the maximum number of features considered for splitting a node, the maximum depth, the minimum number of points placed in a node before the node is split, and the minimum number of points allowed in a leaf node.
To find parameter values, we performed a 10-fold cross-validation, thus we evaluated average values and standard errors of RMSE and $R^2$ (cf. later) on the test sets of such 10 different splits of the data each time.
After finding the parameters, for the sake of simplicity, we analyzed SHAP summary plots on a single data split in 80\% train and 20\% test.
The whole prediction framework was implemented by considering the models, the methods, and the evaluation measures present in scikit-learn\footnote{\url{https://scikit-learn.org/stable/}} and the SHAP library\footnote{\url{https://shap.readthedocs.io/en/latest/}}.

\noindent 
\textbf{Evaluation details}. We evaluate the models with the root-mean-square error (RMSE) and the coefficient of determination ($R^2$).

To introduce RMSE, we first define the sum of the square of errors, or residual sum of squares, $RSS$, as follows:
\begin{equation}
RSS = \sum_i^N{(y_i-\hat{y}_i})^2,
\end{equation}
where $N$ is the number of words, $y_i$ is the empirical concreteness score of a word in the Glasgow Norms, and $\hat{y}_i$ is the score predicted by a model for that word.
To understand this in our context, let us consider a model that predicts, respectively, a concreteness score of $6.5$ and another of $4.5$ for the two words \textit{brain} and \textit{mind}, which have, respectively, empirical ground truth values of $6.4$ and $2.5$ in the Glasgow Norms.
The $RSS$ is of $4.01$, indicating there is, to some extent, some amount of error between the predicted and the empirical values.
To better read the errors, it is often used RMSE, namely the square root of the average of $RSS$. Formally:
\begin{equation}
\centering
RMSE=\sqrt{\frac{1}{N}*RSS}.
\end{equation}
In our toy example, the average of $RSS$ is $2.005$, thus $RMSE=1.41$, indicating there exists variance in the predicted scores with respect to the empirical ground truth values.

Similarly, to describe $R^2$, we first introduce the total sum of squares, $TSS$, as follows:
\begin{equation}
TSS = \sum_i^N{(y_i-\bar{y}})^2,
\end{equation}
where $\bar{y}$ is the average of the empirical ground truth scores, thus
$TSS$ sums over the squared differences between the empirical ground truth values and their average.
$R^2$ is thus defined as follows:
\begin{equation}
R^2=1-\frac{RSS}{TSS}.
\end{equation}
In the example with the two words above, $\bar{y}$ is 4.45, and $TSS$ is 7.6, and $R^2=0.47$.
A different model that would predict a different value of the word \textit{mind}, e.g., 2.8, would decrease $RMSE$ and increase $R2$ for lower residuals.


\section*{Availability of data and materials}
The original free associations analysed during the current study are available from the Small World of Words website: \url{https://smallworldofwords.org/en/project/}.

The node covariates analysed during the current study are available from the Glasgow Norms paper \cite{scott2019glasgow}.

Data preprocessing analysis is available at the following link: \url{https://github.com/dsalvaz/Hypergraph-Cognitive-Networks}.  

\section*{Competing interests}
  The authors declare that they have no competing interests.

\section*{Author's contributions}

Conceptualization: SC, MS and GR; Data curation: SC, SD, MS and GR; Formal analysis: SC, MS and GR; Investigation: All authors; Methodology: SC, MS and GR; Supervision: MS and GR; Validation: SD and MS; Visualization: SC; Roles/Writing - original draft: All authors.

\section*{Acknowledgments}
This work is supported by the European Union – Horizon 2020 Program under the scheme “INFRAIA-01-2018-2019 – Integrating Activities for Advanced Communities”, Grant Agreement n.871042, “SoBigData++: European Integrated Infrastructure for Social Mining and Big Data Analytics” (http://www.sobigdata.eu)  and by the CHIST-ERA grant CHIST-ERA-19-XAI-010, by MUR (grant No. not yet available), FWF (grant No. I 5205), EPSRC (grant No. EP/V055712/1), NCN (grant No. 2020/02/Y/ST6/00064), ETAg (grant No. SLTAT21096), BNSF (grant No. \begin{otherlanguage*}{russian}КП-06-ДОО2/5\end{otherlanguage*}).


\printbibliography

@incollection{bhagat2011node,
  title={Node classification in social networks},
  author={Bhagat, Smriti and Cormode, Graham and Muthukrishnan, S},
  booktitle={Social network data analytics},
  pages={115--148},
  year={2011},
  publisher={Springer}
}

@article{christianson2020architecture,
  title={Architecture and evolution of semantic networks in mathematics texts},
  author={Christianson, Nicolas H and Sizemore Blevins, Ann and Bassett, Danielle S},
  journal={Proceedings of the Royal Society A},
  volume={476},
  number={2239},
  pages={20190741},
  year={2020},
  publisher={The Royal Society Publishing}
}

@article{firth1957synopsis,
  title={A synopsis of linguistic theory, 1930-1955},
  author={Firth, John R},
  journal={Studies in linguistic analysis},
  year={1957},
  publisher={Basil Blackwell}
}

@article{lenci2018distributional,
  title={Distributional models of word meaning},
  author={Lenci, Alessandro},
  journal={Annual review of Linguistics},
  volume={4},
  pages={151--171},
  year={2018},
  publisher={Annual Reviews}
}

@article{fisher1922goodness,
  title={The goodness of fit of regression formulae, and the distribution of regression coefficients},
  author={Fisher, Ronald A},
  journal={Journal of the Royal Statistical Society},
  volume={85},
  number={4},
  pages={597--612},
  year={1922},
  publisher={JSTOR}
}

@article{platt1999probabilistic,
  title={Probabilistic outputs for support vector machines and comparisons to regularized likelihood methods},
  author={Platt, John and others},
  journal={Advances in large margin classifiers},
  volume={10},
  number={3},
  pages={61--74},
  year={1999},
  publisher={Cambridge, MA}
}

@incollection{schapire2013explaining,
  title={Explaining adaboost},
  author={Schapire, Robert E},
  booktitle={Empirical inference},
  pages={37--52},
  year={2013},
  publisher={Springer}
}

@article{freund1997decision,
  title={A decision-theoretic generalization of on-line learning and an application to boosting},
  author={Freund, Yoav and Schapire, Robert E},
  journal={Journal of computer and system sciences},
  volume={55},
  number={1},
  pages={119--139},
  year={1997},
  publisher={Elsevier}
}

@article{breiman2001random,
  title={Random forests},
  author={Breiman, Leo},
  journal={Machine learning},
  volume={45},
  number={1},
  pages={5--32},
  year={2001},
  publisher={Springer}
}

@article{citraro2020identifying,
  title={Identifying and exploiting homogeneous communities in labeled networks},
  author={Citraro, Salvatore and Rossetti, Giulio},
  journal={Applied Network Science},
  volume={5},
  number={1},
  pages={1--20},
  year={2020},
  publisher={SpringerOpen}
}

@inproceedings{whang2013overlapping,
  title={Overlapping community detection using seed set expansion},
  author={Whang, Joyce Jiyoung and Gleich, David F and Dhillon, Inderjit S},
  booktitle={Proceedings of the 22nd ACM international conference on Information \& Knowledge Management},
  pages={2099--2108},
  year={2013}
}

@inproceedings{kloumann2014community,
  title={Community membership identification from small seed sets},
  author={Kloumann, Isabel M and Kleinberg, Jon M},
  booktitle={Proceedings of the 20th ACM SIGKDD international conference on Knowledge discovery and data mining},
  pages={1366--1375},
  year={2014}
}

@article{blondel2008fast,
  title={Fast unfolding of communities in large networks},
  author={Blondel, Vincent D and Guillaume, Jean-Loup and Lambiotte, Renaud and Lefebvre, Etienne},
  journal={Journal of statistical mechanics: theory and experiment},
  volume={2008},
  number={10},
  pages={P10008},
  year={2008},
  publisher={IOP Publishing}
}

@article{zemla2020snafu,
  title={SNAFU: The semantic network and fluency utility},
  author={Zemla, Jeffrey C and Cao, Kesong and Mueller, Kimberly D and Austerweil, Joseph L},
  journal={Behavior research methods},
  volume={52},
  number={4},
  pages={1681--1699},
  year={2020},
  publisher={Springer}
}

@inproceedings{li2015uncovering,
  title={Uncovering the small community structure in large networks: A local spectral approach},
  author={Li, Yixuan and He, Kun and Bindel, David and Hopcroft, John E},
  booktitle={Proceedings of the 24th international conference on world wide web},
  pages={658--668},
  year={2015}
}

@inproceedings{zipf2016human,
  title={Human behavior and the principle of least effort: An introduction to human ecology},
  author={Zipf, George Kingsley},
  year={2016},
  booktitle={Ravenio Books}
}

@article{miller1995wordnet,
  title={WordNet: a lexical database for English},
  author={Miller, George A},
  journal={Communications of the ACM},
  volume={38},
  number={11},
  pages={39--41},
  year={1995},
  publisher={ACM New York, NY, USA}
}

@article{mcpherson2001birds,
  title={Birds of a feather: Homophily in social networks},
  author={McPherson, Miller and Smith-Lovin, Lynn and Cook, James M},
  journal={Annual review of sociology},
  volume={},
  number={},
  pages={},
  year={2001},
  publisher={Annual Reviews 4139 El Camino Way, PO Box 10139, Palo Alto, CA 94303-0139, USA}
}

@article{newman2003mixing,
  title={Mixing patterns in networks},
  author={Newman, Mark EJ},
  journal={Physical review E},
  volume={67},
  number={2},
  pages={026126},
  year={2003},
  publisher={APS}
}

@phdthesis{barbaresi2014language,
  title={Language-classified Open Subtitles (LACLOS): download, extraction, and quality assessment},
  author={Barbaresi, Adrien},
  year={2014},
  school={BBAW}
}

@article{kuperman2012age,
  title={Age-of-acquisition ratings for 30,000 English words},
  author={Kuperman, Victor and Stadthagen-Gonzalez, Hans and Brysbaert, Marc},
  journal={Behavior research methods},
  volume={44},
  number={4},
  pages={978--990},
  year={2012},
  publisher={Springer}
}

@inproceedings{comrie2021hypergraph,
  title={Hypergraph Ego-networks and Their Temporal Evolution},
  author={Comrie, Cazamere and Kleinberg, Jon},
  booktitle={2021 IEEE International Conference on Data Mining (ICDM)},
  pages={91--100},
  year={2021},
  organization={IEEE}
}

@article{siew2013community,
  title={Community structure in the phonological network},
  author={Siew, Cynthia SQ},
  journal={Frontiers in psychology},
  volume={4},
  pages={553},
  year={2013},
  publisher={Frontiers}
}

@article{citraro2022feature,
  title={Feature-rich multiplex lexical networks reveal mental strategies of early language learning},
  author={Citraro, Salvatore and Vitevitch, Michael S and Stella, Massimo and Rossetti, Giulio},
  journal={arXiv preprint arXiv:2201.05061},
  year={2022}
}

@article{van2015examining,
  title={Examining assortativity in the mental lexicon: Evidence from word associations},
  author={Van Rensbergen, Bram and Storms, Gert and De Deyne, Simon},
  journal={Psychonomic bulletin \& review},
  volume={22},
  number={6},
  pages={1717--1724},
  year={2015},
  publisher={Springer}
}

@article{doczi2019overview,
  title={An overview of conceptual models and theories of lexical representation in the mental lexicon},
  author={D{\'o}czi, Brigitta},
  journal={The Routledge Handbook of Vocabulary Studies},
  pages={46--65},
  year={2019},
  publisher={Routledge}
}

@article{lundberg2020local2global,
  title={From local explanations to global understanding with explainable AI for trees},
  author={Lundberg, Scott M. and Erion, Gabriel and Chen, Hugh and DeGrave, Alex and Prutkin, Jordan M. and Nair, Bala and Katz, Ronit and Himmelfarb, Jonathan and Bansal, Nisha and Lee, Su-In},
  journal={Nature Machine Intelligence},
  volume={2},
  number={1},
  pages={2522-5839},
  year={2020},
  publisher={Nature Publishing Group}
}

@inproceedings{NIPS2017_7062,
title = {A Unified Approach to Interpreting Model Predictions},
author = {Lundberg, Scott M and Lee, Su-In},
booktitle = {Advances in Neural Information Processing Systems 30},
editor = {I. Guyon and U. V. Luxburg and S. Bengio and H. Wallach and R. Fergus and S. Vishwanathan and R. Garnett},
pages = {4765--4774},
year = {2017},
booktitle = {Curran Associates, Inc.},
url = {http://papers.nips.cc/paper/7062-a-unified-approach-to-interpreting-model-predictions.pdf}
}

@article{fortunato2016community,
  title={Community detection in networks: A user guide},
  author={Fortunato, Santo and Hric, Darko},
  journal={Physics reports},
  volume={659},
  pages={1--44},
  year={2016},
  publisher={Elsevier}
}

@article{de2019small,
  title={The “Small World of Words” English word association norms for over 12,000 cue words},
  author={De Deyne, Simon and Navarro, Danielle J and Perfors, Amy and Brysbaert, Marc and Storms, Gert},
  journal={Behavior research methods},
  volume={51},
  number={3},
  pages={987--1006},
  year={2019},
  publisher={Springer}
}

@inproceedings{aitchison2012words,
  title={Words in the mind: An introduction to the mental lexicon},
  author={Aitchison, Jean},
  year={2012},
  booktitle={John Wiley \& Sons}
}

@article{montefinese2019semantic,
  title={Semantic representation of abstract and concrete words: A minireview of neural evidence},
  author={Montefinese, Maria},
  journal={Journal of neurophysiology},
  volume={121},
  number={5},
  pages={1585--1587},
  year={2019},
  publisher={American Physiological Society Bethesda, MD}
}

@article{scott2019glasgow,
  title={The Glasgow Norms: Ratings of 5,500 words on nine scales},
  author={Scott, Graham G and Keitel, Anne and Becirspahic, Marc and Yao, Bo and Sereno, Sara C},
  journal={Behavior research methods},
  volume={51},
  number={3},
  pages={1258--1270},
  year={2019},
  publisher={Springer}
}

@article{siew2019cognitive,
  title={Cognitive network science: A review of research on cognition through the lens of network representations, processes, and dynamics},
  author={Siew, Cynthia SQ and Wulff, Dirk U and Beckage, Nicole M and Kenett, Yoed N},
  journal={Complexity},
  volume={2019},
  year={2019},
  publisher={Hindawi}
}

@article{wulff2022using,
  title={Using network science to understand the aging lexicon: Linking individuals' experience, semantic networks, and cognitive performance},
  author={Wulff, Dirk U and De Deyne, Simon and Aeschbach, Samuel and Mata, Rui},
  journal={Topics in Cognitive Science},
  volume={14},
  number={1},
  pages={93--110},
  year={2022},
  publisher={Wiley Online Library}
}

@article{vitevitch2021phonological,
  title={Phonological but not semantic influences on the speech-to-song illusion},
  author={Vitevitch, Michael S and Ng, Joshua W and Hatley, Evan and Castro, Nichol},
  journal={Quarterly Journal of Experimental Psychology},
  volume={74},
  number={4},
  pages={585--597},
  year={2021},
  publisher={SAGE Publications Sage UK: London, England}
}

@article{de2021visual,
  title={Visual and affective multimodal models of word meaning in language and mind},
  author={De Deyne, Simon and Navarro, Danielle J and Collell, Guillem and Perfors, Andrew},
  journal={Cognitive Science},
  volume={45},
  number={1},
  pages={e12922},
  year={2021},
  publisher={Wiley Online Library}
}

@inproceedings{kennington2021enriching,
  title={Enriching language models with visually-grounded word vectors and the Lancaster sensorimotor norms},
  author={Kennington, Casey},
  booktitle={Proceedings of the 25th Conference on Computational Natural Language Learning},
  pages={148--157},
  year={2021}
}

@article{vitevitch2022can,
  title={What can network science tell us about phonology and language processing?},
  author={Vitevitch, Michael S},
  journal={Topics in Cognitive Science},
  volume={14},
  number={1},
  pages={127--142},
  year={2022},
  publisher={Wiley Online Library}
}

@article{zock2010deliberate,
  title={Deliberate word access: an intuition, a roadmap and some preliminary empirical results},
  author={Zock, Michael and Ferret, Olivier and Schwab, Didier},
  journal={International Journal of Speech Technology},
  volume={13},
  number={4},
  pages={201--218},
  year={2010},
  publisher={Springer}
}

@article{steyvers2005large,
  title={The large-scale structure of semantic networks: Statistical analyses and a model of semantic growth},
  author={Steyvers, Mark and Tenenbaum, Joshua B},
  journal={Cognitive science},
  volume={29},
  number={1},
  pages={41--78},
  year={2005},
  publisher={Wiley Online Library}
}

@article{castro2020contributions,
  title={Contributions of modern network science to the cognitive sciences: Revisiting research spirals of representation and process},
  author={Castro, Nichol and Siew, Cynthia SQ},
  journal={Proceedings of the Royal Society A},
  volume={476},
  number={2238},
  pages={20190825},
  year={2020},
  publisher={The Royal Society Publishing}
}

@article{stella2017multiplex,
  title={Multiplex lexical networks reveal patterns in early word acquisition in children},
  author={Stella, Massimo and Beckage, Nicole M and Brede, Markus},
  journal={Scientific reports},
  volume={7},
  number={1},
  pages={1--10},
  year={2017},
  publisher={Nature Publishing Group}
}

@article{de2013better,
  title={Better explanations of lexical and semantic cognition using networks derived from continued rather than single-word associations},
  author={De Deyne, Simon and Navarro, Daniel J and Storms, Gert},
  journal={Behavior research methods},
  volume={45},
  number={2},
  pages={480--498},
  year={2013},
  publisher={Springer}
}

@article{kenett2014investigating,
  title={Investigating the structure of semantic networks in low and high creative persons},
  author={Kenett, Yoed N and Anaki, David and Faust, Miriam},
  journal={Frontiers in human neuroscience},
  volume={8},
  pages={407},
  year={2014},
  publisher={Frontiers Media SA}
}

@article{stella2019forma,
  title={Forma mentis networks quantify crucial differences in STEM perception between students and experts},
  author={Stella, Massimo and De Nigris, Sarah and Aloric, Aleksandra and Siew, Cynthia SQ},
  journal={PloS one},
  volume={14},
  number={10},
  pages={e0222870},
  year={2019},
  publisher={Public Library of Science San Francisco, CA USA}
}

@article{stella2019viability,
  title={Viability in multiplex lexical networks and machine learning characterizes human creativity},
  author={Stella, Massimo and Kenett, Yoed N},
  journal={Big Data and Cognitive Computing},
  volume={3},
  number={3},
  pages={45},
  year={2019},
  publisher={MDPI}
}

@article{fatima2021dasentimental,
  title={DASentimental: Detecting Depression, Anxiety, and Stress in Texts via Emotional Recall, Cognitive Networks, and Machine Learning},
  author={Fatima, Asra and Li, Ying and Hills, Thomas Trenholm and Stella, Massimo},
  journal={Big Data and Cognitive Computing},
  volume={5},
  number={4},
  pages={77},
  year={2021},
  publisher={MDPI}
}

@article{hills2015foraging,
  title={Foraging in semantic fields: How we search through memory},
  author={Hills, Thomas T and Todd, Peter M and Jones, Michael N},
  journal={Topics in cognitive science},
  volume={7},
  number={3},
  pages={513--534},
  year={2015},
  publisher={Wiley Online Library}
}

@article{brysbaert2017test,
  title={Test-based age-of-acquisition norms for 44 thousand English word meanings},
  author={Brysbaert, Marc and Biemiller, Andrew},
  journal={Behavior research methods},
  volume={49},
  number={4},
  pages={1520--1523},
  year={2017},
  publisher={Springer}
}

@article{todd2020foraging,
  title={Foraging in mind},
  author={Todd, Peter M and Hills, Thomas T},
  journal={Current Directions in Psychological Science},
  volume={29},
  number={3},
  pages={309--315},
  year={2020},
  publisher={Sage Publications Sage CA: Los Angeles, CA}
}

@article{vankrunkelsven2018predicting,
  title={Predicting lexical norms: A comparison between a word association model and text-based word co-occurrence models},
  author={Vankrunkelsven, Hendrik and Verheyen, Steven and Storms, Gert and De Deyne, Simon},
  journal={Journal of cognition},
  volume={1},
  number={1},
  year={2018},
  publisher={Ubiquity Press}
}

@inproceedings{de2015using,
  title={Using network clustering to uncover the taxonomic and thematic structure of the mental lexicon},
  author={De Deyne, Simon and Verheyen, Steven},
  booktitle={CEUR Workshop Proceedings},
  volume={1347},
  pages={172--176},
  year={2015}
}

@article{veldt2023combinatorial,
  title={Combinatorial characterizations and impossibilities for higher-order homophily},
  author={Veldt, Nate and Benson, Austin R and Kleinberg, Jon},
  journal={Science Advances},
  volume={9},
  number={1},
  pages={eabq3200},
  year={2023},
  publisher={American Association for the Advancement of Science}
}

@article{musciotto2021detecting,
  title={Detecting informative higher-order interactions in statistically validated hypergraphs},
  author={Musciotto, Federico and Battiston, Federico and Mantegna, Rosario N},
  journal={Communications Physics},
  volume={4},
  number={1},
  pages={1--9},
  year={2021},
  publisher={Nature Publishing Group}
}

@article{ditzfeld2014self,
  title={Self-structure and emotional experience},
  author={Ditzfeld, Christopher P and Showers, Carolin J},
  journal={Cognition \& emotion},
  volume={28},
  number={4},
  pages={596--621},
  year={2014},
  publisher={Taylor \& Francis}
}

@article{battiston2021physics,
  title={The physics of higher-order interactions in complex systems},
  author={Battiston, Federico and Amico, Enrico and Barrat, Alain and Bianconi, Ginestra and Ferraz de Arruda, Guilherme and Franceschiello, Benedetta and Iacopini, Iacopo and K{\'e}fi, Sonia and Latora, Vito and Moreno, Yamir and others},
  journal={Nature Physics},
  volume={17},
  number={10},
  pages={1093--1098},
  year={2021},
  publisher={Nature Publishing Group}
}

@article{rosas2022disentangling,
  title={Disentangling high-order mechanisms and high-order behaviours in complex systems},
  author={Rosas, Fernando E and Mediano, Pedro AM and Luppi, Andrea I and Varley, Thomas F and Lizier, Joseph T and Stramaglia, Sebastiano and Jensen, Henrik J and Marinazzo, Daniele},
  journal={Nature Physics},
  volume={18},
  number={5},
  pages={476--477},
  year={2022},
  publisher={Nature Publishing Group}
}

@inproceedings{berge1984hypergraphs,
  title={Hypergraphs: combinatorics of finite sets},
  author={Berge, Claude},
  volume={45},
  year={1984},
  booktitle={Elsevier}
}

@article{marinazzo2022information,
  title={An information-theoretic approach to hypergraph psychometrics},
  author={Marinazzo, Daniele and Van Roozendaal, Jan and Rosas, Fernando E and Stella, Massimo and Comolatti, Renzo and Colenbier, Nigel and Stramaglia, Sebastiano and Rosseel, Yves},
  journal={arXiv preprint arXiv:2205.01035},
  year={2022}
}

@misc{battiston2022higher,
  title={Higher-Order Systems},
  author={Battiston, Federico and Petri, Giovanni},
  publisher={Springer}
}

@book{rogers2004semantic,
  title={Semantic cognition: A parallel distributed processing approach},
  author={Rogers, Timothy T and McClelland, James L and others},
  year={2004},
  publisher={MIT press}
}

@article{farah1991computational,
  title={A computational model of semantic memory impairment: modality specificity and emergent category specificity.},
  author={Farah, Martha J and McClelland, James L},
  journal={Journal of experimental psychology: General},
  volume={120},
  number={4},
  pages={339},
  year={1991},
  publisher={American Psychological Association}
}

@article{schapiro2017complementary,
  title={Complementary learning systems within the hippocampus: a neural network modelling approach to reconciling episodic memory with statistical learning},
  author={Schapiro, Anna C and Turk-Browne, Nicholas B and Botvinick, Matthew M and Norman, Kenneth A},
  journal={Philosophical Transactions of the Royal Society B: Biological Sciences},
  volume={372},
  number={1711},
  pages={20160049},
  year={2017},
  publisher={The Royal Society}
}

@article{shabahang2022generalization,
  title={Generalization at retrieval using associative networks with transient weight changes},
  author={Shabahang, Kevin D and Yim, Hyungwook and Dennis, Simon J},
  journal={Computational Brain \& Behavior},
  volume={5},
  number={1},
  pages={124--155},
  year={2022},
  publisher={Springer}
}

@article{gao2022scope,
  title={Scope: The south carolina psycholinguistic metabase},
  author={Gao, Chuanji and Shinkareva, Svetlana V and Desai, Rutvik H},
  journal={Behavior Research Methods},
  pages={1--32},
  year={2022},
  publisher={Springer}
}

@article{brysbaert2000age,
  title={Age-of-acquisition effects in semantic processing tasks},
  author={Brysbaert, Marc and Van Wijnendaele, Ilse and De Deyne, Simon},
  journal={Acta psychologica},
  volume={104},
  number={2},
  pages={215--226},
  year={2000},
  publisher={Elsevier}
}

@inproceedings{charbonnier2019predicting,
  title={Predicting word concreteness and imagery},
  author={Charbonnier, Jean and Wartena, Christian},
  booktitle={Proceedings of the 13th International Conference on Computational Semantics-Long Papers},
  pages={176--187},
  year={2019},
  organization={Association for Computational Linguistics}
}

@article{de2020social,
  title={Social contagion models on hypergraphs},
  author={de Arruda, Guilherme Ferraz and Petri, Giovanni and Moreno, Yamir},
  journal={Physical Review Research},
  volume={2},
  number={2},
  pages={023032},
  year={2020},
  publisher={APS}
}

@article{yao2013semantic,
  title={Semantic size of abstract concepts: It gets emotional when you can’t see it},
  author={Yao, Bo and Vasiljevic, Milica and Weick, Mario and Sereno, Margaret E and O’Donnell, Patrick J and Sereno, Sara C},
  journal={Plos one},
  volume={8},
  number={9},
  pages={e75000},
  year={2013},
  publisher={Public Library of Science San Francisco, USA}
}

@inproceedings{sarker2023generalizing,
  title={Generalizing homophily to simplicial complexes},
  author={Sarker, Arnab and Northrup, Natalie and Jadbabaie, Ali},
  booktitle={Complex Networks and Their Applications XI: Proceedings of The Eleventh International Conference on Complex Networks and their Applications: COMPLEX NETWORKS 2022—Volume 2},
  pages={311--323},
  year={2023},
  organization={Springer}
}

@article{kumar2020distant,
  title={Distant connectivity and multiple-step priming in large-scale semantic networks.},
  author={Kumar, Abhilasha A and Balota, David A and Steyvers, Mark},
  journal={Journal of Experimental Psychology: Learning, Memory, and Cognition},
  volume={46},
  number={12},
  pages={2261},
  year={2020},
  publisher={American Psychological Association}
}

@inproceedings{kumar2020problems,
  title={Problems with Shapley-value-based explanations as feature importance measures},
  author={Kumar, I Elizabeth and Venkatasubramanian, Suresh and Scheidegger, Carlos and Friedler, Sorelle},
  booktitle={International Conference on Machine Learning},
  pages={5491--5500},
  year={2020},
  organization={PMLR}
}

@article{rudin2019stop,
  title={Stop explaining black box machine learning models for high stakes decisions and use interpretable models instead},
  author={Rudin, Cynthia},
  journal={Nature Machine Intelligence},
  volume={1},
  number={5},
  pages={206--215},
  year={2019},
  publisher={Nature Publishing Group}
}

@inproceedings{ghorbani2019data,
  title={Data shapley: Equitable valuation of data for machine learning},
  author={Ghorbani, Amirata and Zou, James},
  booktitle={International Conference on Machine Learning},
  pages={2242--2251},
  year={2019},
  organization={PMLR}
}

@article{fliessbach2006effect,
  title={The effect of word concreteness on recognition memory},
  author={Fliessbach, Klaus and Weis, Susanne and Klaver, Peter and Elger, Christian Erich and Weber, Bernd},
  journal={NeuroImage},
  volume={32},
  number={3},
  pages={1413--1421},
  year={2006},
  publisher={Elsevier}
}

@article{brysbaert2014concreteness,
  title={Concreteness ratings for 40 thousand generally known English word lemmas},
  author={Brysbaert, Marc and Warriner, Amy Beth and Kuperman, Victor},
  journal={Behavior research methods},
  volume={46},
  number={3},
  pages={904--911},
  year={2014},
  publisher={Springer}
}

@article{battiston2020networks,
  title={Networks beyond pairwise interactions: structure and dynamics},
  author={Battiston, Federico and Cencetti, Giulia and Iacopini, Iacopo and Latora, Vito and Lucas, Maxime and Patania, Alice and Young, Jean-Gabriel and Petri, Giovanni},
  journal={Physics Reports},
  volume={874},
  pages={1--92},
  year={2020},
  publisher={Elsevier}
}

@article{kenett2017semantic,
  title={The semantic distance task: Quantifying semantic distance with semantic network path length.},
  author={Kenett, Yoed N and Levi, Effi and Anaki, David and Faust, Miriam},
  journal={Journal of Experimental Psychology: Learning, Memory, and Cognition},
  volume={43},
  number={9},
  pages={1470},
  year={2017},
  publisher={American Psychological Association}
}

@article{valba2022k,
  title={K-clique percolation in free association networks and the possible mechanism behind the  $7\pm 2$ law},
  author={Valba, Olga and Gorsky, Alexander},
  journal={Scientific reports},
  volume={12},
  number={1},
  pages={1--9},
  year={2022},
  publisher={Nature Publishing Group}
}

\pagebreak

\begin{appendices}

\renewcommand{\thetable}{\Alph{section}\arabic{table}}
\renewcommand{\thefigure}{\Alph{section}\arabic{figure}}    
\setcounter{figure}{0}    

\section{Gap in hyperedges}

\begin{table}[]
\centering
\caption{Random Forest evaluation of concreteness prediction based on the Lemon and Hypergraph aggregation strategies with gaps, i.e., without the target word within the contexts.}
\label{tab:alg_gap}
\resizebox{0.6\textwidth}{!}{%
\begin{tabular}{|c|c|c|c|}
\hline
                      &     & G: Lemon (with gap) & Hypergraph (with gap) \\ \hline
\multirow{2}{*}{RMSE} & M   & 1.11                & 1.08                  \\ \cline{2-4} 
                      & SE & 0.04                & 0.03                  \\ \hline
\multirow{2}{*}{$R^2$}   & M   & 0.39                & 0.43                  \\ \cline{2-4} 
                      & SE & 0.04                & 0.05                  \\ \hline
\end{tabular}%
}
\end{table}

An important point to discuss is the question whether including or not including the target word within the context in the hyperedge as well as in the local community obtained with the Lemon algorithm.
We test such a choice within our machine learning framework in predicting concreteness, showing in Table \ref{tab:alg_gap} a decrease in the Random Forest performances on the Lemon- and hypergraph-based sets of features, where the target words are removed from their own contexts, simulating some kind of knowledge gap \cite{christianson2020architecture} in the memory recall patterns.

\section{Performances of other models}

\begin{table}[]
\centering
\caption{Evaluation of concreteness prediction by different regression algorithms on the different sets of aggregation strategies.}
\label{tab:other_alg}
\resizebox{\textwidth}{!}{%
\begin{tabular}{|c|c|c|c|c|c|c|c|c|}
\hline
                                        &                       &   & Non-Net & G: Ego-Net & G: Louvain & G: EVA & G: Lemon & Hypergraph \\ \hline
\multirow{4}{*}{Linear Regression}      & \multirow{2}{*}{RMSE} & M & 1.17    & 1.09       & 1.45       & 1.45   & 1.18     & 1.08       \\ \cline{3-9} 
 &                     & SE & 0.04 & 0.04 & 0.04 & 0.04 & 0.04 & 0.04 \\ \cline{2-9} 
 & \multirow{2}{*}{$R^2$} & M   & 0.33 & 0.42 & 0.02 & 0.03 & 0.31 & 0.44 \\ \cline{3-9} 
 &                     & SE & 0.05 & 0.06 & 0.04 & 0.04 & 0.05 & 0.05 \\ \hline
\multirow{4}{*}{Support Vector Machine} & \multirow{2}{*}{RMSE} & M & 0.98    & 0.98       & 1.44       & 1.43   & 1.03     & 0.93       \\ \cline{3-9} 
 &                     & SE & 0.03 & 0.04 & 0.04 & 0.04 & 0.03 & 0.03 \\ \cline{2-9} 
 & \multirow{2}{*}{$R^2$} & M   & 0.53 & 0.53 & 0.06 & 0.05 & 0.48 & 0.58 \\ \cline{3-9} 
 &                     & SE & 0.04 & 0.06 & 0.03 & 0.03 & 0.05 & 0.04 \\ \hline
\multirow{4}{*}{AdaBoost}               & \multirow{2}{*}{RMSE} & M & 1.13    & 1.10       & 1.45       & 1.45   & 1.16     & 1.06       \\ \cline{3-9} 
 &                     & SE & 0.03 & 0.02 & 0.04 & 0.05 & 0.03 & 0.03 \\ \cline{2-9} 
 & \multirow{2}{*}{$R^2$} & M   & 0.39 & 0.41 & 0.03 & 0.03 & 0.36 & 0.49 \\ \cline{3-9} 
 &                     & SE & 0.03 & 0.04 & 0.03 & 0.04 & 0.03 & 0.03 \\ \hline
\end{tabular}%
}
\end{table}

As highlighted in the main text, the Random Forest predictor on the several different sets of features demonstrated that the hypergraph model achieves better results than the other aggregating strategies.
To ensure that the result does not depend on a specific instance of a particular regressor, in Table \ref{tab:other_alg} we show the performances of other predictors on the same sets of features.
We perform a linear regression, as well as a Support Vector Machine model, and an ensemble method similar to the Random Forest framework but based on boosting.
All the machine learning algorithms provide similar results such that the features based on the hypergraph aggregation continues to provide better performances in terms of RMSE and $R^2$.
The only difference is in the magnitude of the scores, such that the Random Forest performances, presented in the main article, are the highest among all the four regressors.

\section{Predicting other features}

\begin{figure}[t!]
\centering
  \subfloat[Valence]{\includegraphics[scale=0.27]{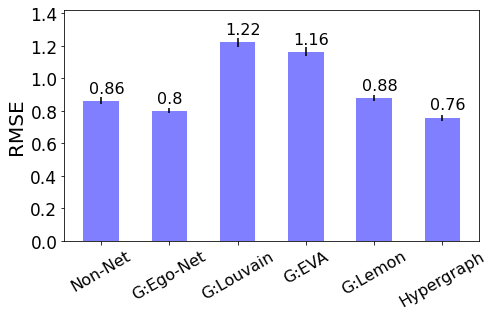}}
  \subfloat[Arousal]{\includegraphics[scale=0.27]{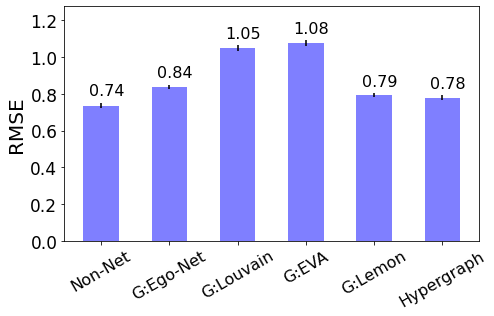}}
  \subfloat[Dominance]{\includegraphics[scale=0.27]{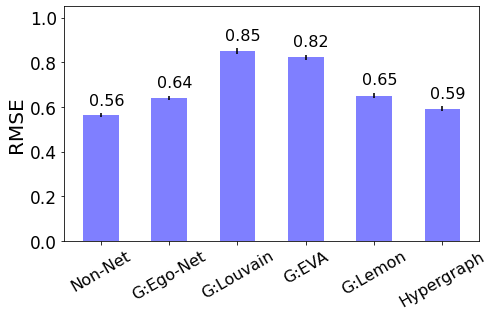}}
  \qquad
  \subfloat[AoA]{\includegraphics[scale=0.27]{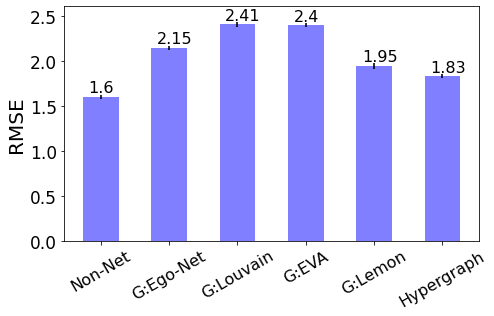}}
  \subfloat[Familiarity]{\includegraphics[scale=0.27]{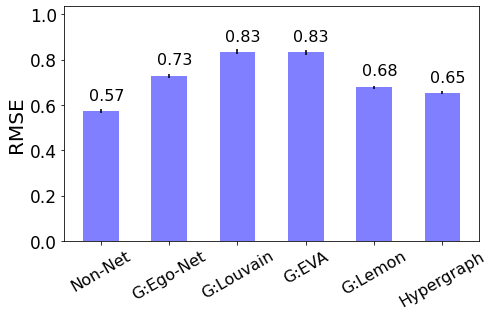}}
  \subfloat[Length]{\includegraphics[scale=0.27]{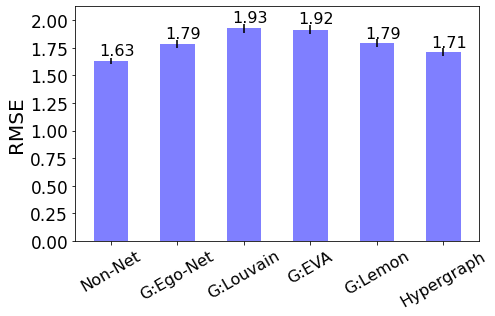}}

  \caption{RMSE -- Random Forest evaluation of several features prediction based on the different aggregation strategies.}
  \label{fig:eval_rf_all_rmse}
      \end{figure}

\begin{figure}[t!]
      \centering
  \subfloat[Valence]{\includegraphics[scale=0.27]{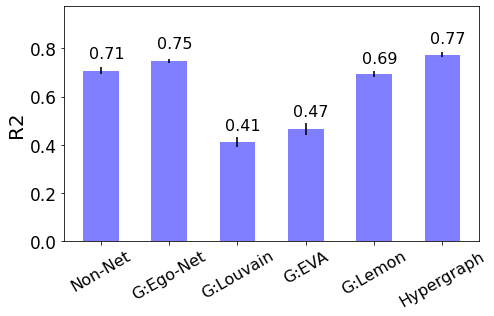}}
  \subfloat[Arousal]{\includegraphics[scale=0.27]{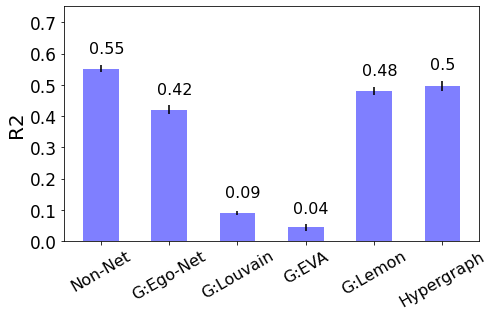}}
  \subfloat[Dominance]{\includegraphics[scale=0.27]{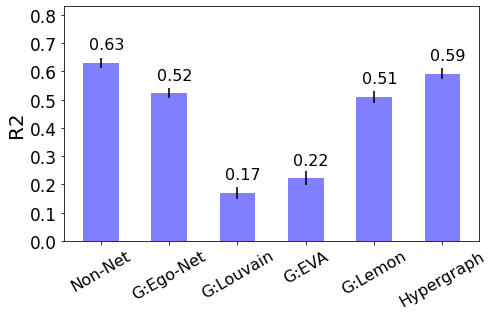}}
  \qquad
  \subfloat[AoA]{\includegraphics[scale=0.27]{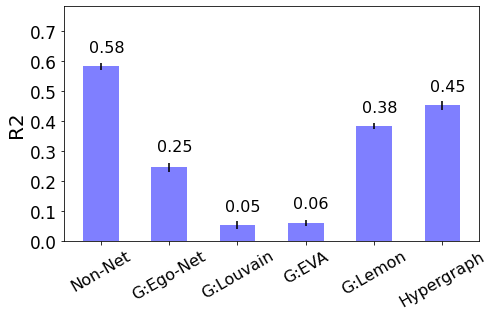}}
  \subfloat[Familiarity]{\includegraphics[scale=0.27]{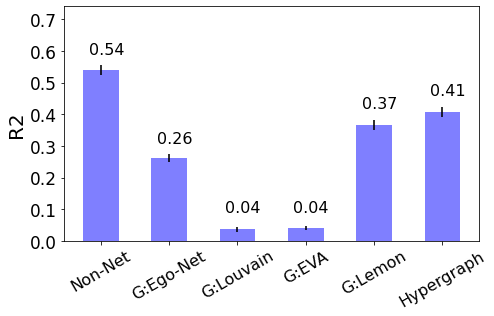}}
  \subfloat[Length]{\includegraphics[scale=0.27]{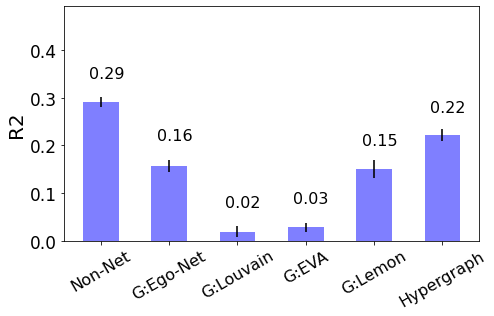}}

  \caption{$R^2$ -- Random Forest evaluation of several features prediction based on the different aggregation strategies.}
  \label{fig:eval_rf_all_r2}
      \end{figure}

As a main research subject for questioning network-based models of human memory, we limited our analysis in predicting concept concreteness.
Figure \ref{fig:eval_rf_all_rmse} and Figure \ref{fig:eval_rf_all_r2} highlight a supplemental analysis, and show the results for the prediction of other features.
Again, we compare the hypergraph strategy against the other graph-based and empirical representations already described in the main work.
The same methodology for regression is applied as well, i.e., a hyperparameter-tuned Random Forest.
We choose to compare the dimensions of valence, arousal, dominance, age of acquisition, familiarity and length, expecting different performances for them across the several aggregation strategies.
Results tell us that, similarly to what we observed with concreteness, a hypergraph aggregation strategy leads to better estimate valence, while the empirical values let the model perform better for all the other dimensions.
As discussed in the main text, non-semantic psycholinguistic features might not give rise to compartmentalisation, as we particularly observe for AoA, familiarity, and length.

\section{Other aggregation strategies}

\begin{table}[]
\centering
\caption{Random Forest evaluation of concreteness prediction based on alternative pairwise network constructions.}
\label{tab:other_graph}
\begin{tabular}{|cccccc|}
\hline
\multicolumn{6}{|c|}{G: Ego-Net}                                                                                                                  \\ \hline
\multicolumn{1}{|l|}{} & \multicolumn{1}{l|}{}    & \multicolumn{2}{l|}{Edges btw resp.}                  & \multicolumn{2}{l|}{Edges btw. resp.} \\ \hline
\multicolumn{1}{|c|}{} & \multicolumn{1}{c|}{}    & \multicolumn{1}{c|}{R1}   & \multicolumn{1}{c|}{R123} & \multicolumn{1}{c|}{Chain}  & Clique  \\ \hline
\multicolumn{1}{|c|}{\multirow{2}{*}{RMSE}} & \multicolumn{1}{c|}{M} & \multicolumn{1}{c|}{1.03} & \multicolumn{1}{c|}{1.01} & \multicolumn{1}{c|}{0.95} & 0.94 \\ \cline{2-6} 
\multicolumn{1}{|c|}{} & \multicolumn{1}{c|}{STD} & \multicolumn{1}{c|}{0.03} & \multicolumn{1}{c|}{0.02} & \multicolumn{1}{c|}{0.03}   & 0.05    \\ \hline
\multicolumn{1}{|c|}{\multirow{2}{*}{R2}}   & \multicolumn{1}{c|}{M} & \multicolumn{1}{c|}{0.48} & \multicolumn{1}{c|}{0.50} & \multicolumn{1}{c|}{0.54} & 0.54 \\ \cline{2-6} 
\multicolumn{1}{|c|}{} & \multicolumn{1}{c|}{STD} & \multicolumn{1}{c|}{0.05} & \multicolumn{1}{c|}{0.02} & \multicolumn{1}{c|}{0.05}   & 0.05    \\ \hline
\end{tabular}
\end{table}

In this work, we tried to cover all the fundamental network-based aggregation strategies among pairwise ego-networks, graph communities and high-order ego-network representations, aiming to re-elaborate the features' values of a target word.
However, other aggregation strategies may come to mind and, consequently, they may affect the results of a prediction.
For instance, regarding the graph ego-network strategy, several other options are possible.
In the main text, we represented the pairwise network using the so-called \textit{R123} strategy, where links are placed between the cue word and the three responses, without connecting in their turn the responses (cf. \textit{Materials and Methods, Aggregation details}).
However, one might think that this strategy gives more importance to the cue word than to the responses.
To validate the pairwise ego-network strategy, we also implemented other variants, particularly:
\begin{itemize}
\item the more straightforward \textit{R1}, where the cue word is connected only to the first response;
\item a variant where links are placed following a \textit{chain}, e.g., the cue word is linked to the first response, then the second response is linked to the second response, etc;
\item (iii) a variant where the cue word is linked to the three responses, and all the responses are in their turn connected to each other.
\end{itemize}
The last variant, in particular, can be thought of as another hypergraph-based strategy rather than a pairwise graph-based one, since each free association is represented as a clique.
Also, we can distinguish the strategies according to the fact that some of them (R1 and R123) place edges between the cue word and the responses only, while other ones (chain- and clique-based) include edges between the responses as well, a procedure that gives more importance to the whole group.

The resulting graph $G_{R1}=(V_G,E_G)$, with the filtering due to the matching between the SWoW and the Glasgow Norms words, has $V_G=3581$ and $E_G=61,359$. Similarly, $G_{Chain}=(V_G,E_G)$ has $V_G=3586$ and $E_G=260,104$, and $G_{Clique}=(V_G,E_G)$ has $V_G=3586$ and $E_G=396,573$.
Results are visible in Table \ref{tab:other_graph}. Note that values for R123 are the same presented in the main text.
When only pairwise links between the cue word and the other responses are present (i.e., R1 and R123), results about concreteness prediction seem to be worse, while the performances improve when connections between responses are involved.
These results suggest that, when connections between ``implicit''/``indirect'' words are placed, performances are better, a result that leads to consider the importance of compartmentalised models of free associations.

\end{appendices}

\end{document}